\documentclass{article}

\usepackage{url}
\usepackage[T1]{fontenc}    
\usepackage{amsfonts}       
\usepackage{nicefrac}       
\usepackage{microtype}      
\usepackage{color}
\usepackage{latexsym}
\usepackage{amsmath}
\usepackage{amsthm}
\usepackage{amssymb}
\usepackage{algorithm}
\usepackage{algorithmic}
\usepackage{graphics} 
\usepackage{epsfig} 
\usepackage{wrapfig}

\usepackage{enumitem}
\usepackage{multirow}
\usepackage{array}
\usepackage{bbding}
\usepackage{threeparttable}
\usepackage{thmtools, thm-restate}

\usepackage{xcolor}
\usepackage{graphicx,subfigure}
\usepackage{tcolorbox}

\usepackage{booktabs}
\usepackage{makecell}
\usepackage{bigstrut,rotating}
\usepackage{soul}
\usepackage[textsize=tiny]{todonotes}
\usepackage{pifont}

\usepackage{hyperref}

\usepackage[accepted]{icml2024}

\usepackage{amsmath}
\usepackage{amssymb}
\usepackage{mathtools}
\usepackage{amsthm}

\usepackage[capitalize,noabbrev]{cleveref}

\usepackage[textsize=tiny]{todonotes}

\icmltitlerunning{Understanding Server-Assisted Federated Learning in the Presence of Incomplete Client Participation}

\newcommand{\x}{\mathbf{x}}

\newcommand{\y}{\mathbf{y}}
\newcommand{\z}{\mathbf{z}}

\newcommand{\mc}[1]{\mathcal{#1}}
\newcommand{\mb}[1]{\mathbb{#1}}

\newtheorem{defn}{Definition}

\newtheorem{rem}{Remark}

\newtheorem{assum}{Assumption}

\usepackage{amsmath,amsfonts,bm}

\def\eqref#1{equation~\ref{#1}}

\def\1{\bm{1}}

\DeclareMathAlphabet{\mathsfit}{\encodingdefault}{\sfdefault}{m}{sl}
\SetMathAlphabet{\mathsfit}{bold}{\encodingdefault}{\sfdefault}{bx}{n}

\newcommand{\alg}{$\mathsf{SAFARI}$~}

\begin{document}

\twocolumn[
\icmltitle{Understanding Server-Assisted Federated Learning in the Presence of Incomplete Client Participation}



\icmlsetsymbol{equal}{*}

\begin{icmlauthorlist}
\icmlauthor{Haibo Yang}{rit}
\icmlauthor{Peiwen Qiu}{osu}
\icmlauthor{Prashant Khanduri}{wsu}
\icmlauthor{Minghong Fang}{du}
\icmlauthor{Jia Liu}{osu}
\end{icmlauthorlist}

\icmlaffiliation{rit}{Department of Computing and Information Sciences Ph.D., Rochester Institute of Technology, Rochester, NY, USA}
\icmlaffiliation{osu}{Department of Electrical and Computer Engineering, The Ohio
State University, Columbus, OH, USA}
\icmlaffiliation{wsu}{Department of Computer Science, Wayne State University, Detroit, MI, USA}
\icmlaffiliation{du}{Department of Electrical and Computer Engineering, Duke University, Durham, NC, USA}

\icmlcorrespondingauthor{Haibo Yang}{hbycis@rit.edu}

\icmlkeywords{Learning Theory, Federated Learning}

\vskip 0.3in
]



\printAffiliationsAndNotice{}  


\begin{abstract}
Existing works in federated learning (FL) often assume an ideal system with either full client or uniformly distributed client participation. 
However, in practice, it has been observed that some clients may never participate in FL training (aka incomplete client participation) due to a myriad of system heterogeneity factors.
A popular approach to mitigate impacts of incomplete client participation is the server-assisted federated learning (SA-FL) framework, where the server is equipped with an auxiliary dataset.
However, despite SA-FL has been empirically shown to be effective in addressing the incomplete client participation problem, there remains a lack of theoretical understanding for SA-FL.
Meanwhile, the ramifications of incomplete client participation in conventional FL are also poorly understood.
These theoretical gaps motivate us to rigorously investigate SA-FL.
Toward this end, we first show that conventional FL is {\em not} PAC-learnable under incomplete client participation in the worst case.
Then, we show that the PAC-learnability of FL with incomplete client participation can indeed be revived by SA-FL, which theoretically justifies the use of SA-FL for the first time.
Lastly, to provide practical guidance for SA-FL training under {\em incomplete client participation}, we propose the \alg (\ul{s}erver-\ul{a}ssisted \ul{f}ederated \ul{a}ve\ul{r}ag\ul{i}ng) algorithm that enjoys the same linear convergence speedup guarantees as classic FL with ideal client participation assumptions, offering the first SA-FL algorithm with convergence guarantee.
Extensive experiments on different datasets show \alg significantly improves the performance under incomplete client participation.
\end{abstract}

\section{Introduction} \label{sec: Introduction}

Since the seminal work by~\citet{mcmahan2017communication}, federated learning (FL) has emerged as a powerful distributed learning paradigm that enables a large number of clients (e.g., edge devices) to collaboratively train a model under a central server's coordination.  
However, as FL gaining popularity, it has also become apparent that FL faces a key challenge unseen in traditional distributed learning in data-center settings -- system heterogeneity.
Generally speaking, system heterogeneity in FL is caused by the massively different computation and communication capabilities at each client (computational power, communication capacity, drop-out rate, etc.).
Studies have shown that system heterogeneity can significantly impact client participation in a highly non-trivial fashion and render {\em incomplete client participation}, which severely degrades the learning performance~\cite{bonawitz2019fldesign,yang2021characterizing}.
For example, it is shown in \cite{yang2021characterizing} that more than $30\%$ clients never participate in FL, while only $30\%$ of the clients contribute to $81\%$ of the total computation even if the server uniformly samples the clients.
Exacerbating the problem is the fact that clients' status could be unstable and time-varying due to the aforementioned computation/communication constraints.
This situation sharply contrasts with existing works on FL with partial client participation, which often assume that clients engage based on a known random process~\cite{Karimireddy2020SCAFFOLD,malinovsky2023federated,cho2023convergence}.

To mitigate the impact of incomplete client participation, 
one approach called {\em server-assisted federated learning} (SA-FL) has been widely adopted in real-world FL systems in recent years~(see, e.g.,~\cite{zhao2018niid,wang2021addressing}).
The basic idea of SA-FL is to equip the server with a small auxiliary dataset sampled from population distribution, so that the distribution deviation induced by incomplete client participation can be corrected.
Nonetheless, while SA-FL has empirically demonstrated its considerable efficacy in addressing incomplete client participation problem in practice, there remains {\em a lack of theoretical understanding} for SA-FL.
This motivates us to rigorously investigate the efficacy of SA-FL in the presence of incomplete client participation.

Somewhat counterintuitively, to understand SA-FL, one must first fully understand the impact of incomplete client participation on conventional FL.
In other words, we need to first answer the following fundamental question: 

\begin{tcolorbox}[left=1.2pt,right=1.2pt,top=1.2pt,bottom=1.2pt]
\textbf{(Q1)}: What are the impacts of incomplete client participation on conventional FL learning performance?
\end{tcolorbox}

Upon answering this question, the next important follow-up question regarding SA-FL is: 

\begin{tcolorbox}[left=1.2pt,right=1.2pt,top=1.2pt,bottom=1.2pt]
\textbf{(Q2)}: What benefits could SA-FL bring and how could we theoretically characterize them?
\end{tcolorbox}

Also, just knowing the benefits of SA-FL is not sufficient to provide guidelines on how to use server-side data in designing training algorithms with convergence guarantees.
Therefore, our third fundamental question for SA-FL is: 

\begin{tcolorbox}[left=1.2pt,right=1.2pt,top=1.2pt,bottom=1.2pt]
\textbf{(Q3)}: Is it possible to develop SA-FL training algorithms with provable convergence rates that can match the state-of-the-art rates in conventional FL?
\end{tcolorbox}

Answering these three questions constitutes the rest of this paper, where we address Q1 and Q2 through the lens of PAC (probably approximately correct) learning, while resolving Q3 by proposing a provably convergent SA-FL algorithm.
Our major contributions are summarized as follows:

\vspace{-.1in}
\begin{list}{\labelitemi}{\leftmargin=1em \itemindent=-0.09em \itemsep=.0em}
\item By establishing a {\em worst-case} generalization error lower bound, we rigorously show that classic FL is {\em not} PAC-learnable under incomplete client participation.
In other words, no learning algorithm can approach zero generalization error with incomplete client participation for classic FL even in the limit of infinitely many data samples. 
This insight, though being negative, warrants the necessity of developing new algorithmic techniques and system architectures (e.g., SA-FL) to modify the classic FL framework to mitigate incomplete client participation.

%
\item We prove a new generalization error bound to show that SA-FL can indeed {\em revive the PAC learnability of FL} with incomplete client participation.
We note that this bound could reach zero asymptotically as the number data samples increases.
This is much stronger than previous results in domain adaptation with non-vanishing small error (see Section \ref{sec: RelatedWork} for details).

\item To ensure that SA-FL is provably convergent in training, we propose a new training algorithm for SA-FL called \alg (\ul{s}erver-\ul{a}ssisted \ul{f}ederated \ul{a}ve\ul{r}ag\ul{i}ng).
By carefully designing the server-client update coordination, we show that \alg achieves an $\mathcal{O}(1/\sqrt{mkR})$ convergence rate in non-convex functions and $\tilde{\mathcal{O}}(\frac{1}{R})$ in strongly-convex functions, matching the convergence rates of state-of-the-art classic FL algorithms~\cite{Li2020convergence,yang2021linearspeedup}. 
We also conduct extensive experiments to demonstrate the effectiveness of our \alg algorithm.
\end{list}


\section{Related Work} \label{sec: RelatedWork}

In this section, we provide an overview on two lines of closely related research, namely (i) FL with partial client participation and (ii) domain adaptation.

\textbf{1) Partial Client Participation in Federated Learning:}
Since The seminal FedAvg algorithm~\cite{mcmahan2017communication}, there have been many follow-ups (e.g., ~\cite{Li2020fedprox,wang2020fednova,zhang2020fedpd,acar2021feddyn,Karimireddy2020SCAFFOLD,luo2021nofear,mitra2021linear,karimireddy2021mime,khanduri2021stem,murata2021bias,Dmitrii21arbitrary,yang2021linearspeedup,grudzien2023can,condat2023tamuna,mishchenko2022proxskip} and so on) on addressing the data heterogeneity challenge in FL.
However, most of these works are based on the full or uniform (i.e., sampling clients uniformly at random) client participation assumption.
%

A related line of works in FL different from full/uniform client participation focuses on {\em proactively creating} flexible client participation (see, e.g.,~\cite{xie2019asynchronous,ruan2021flexible,gu2021fast,Dmitrii21arbitrary,yang2022anarchic,wang2022unified,koloskova2022sharper}).
The main idea here is to allow asynchronous communication or fixed participation pattern (e.g., given probability) for clients to flexibly participate in training.
Existing works in this area often require extra assumptions, such as bounded delay~\cite{ruan2021flexible,gu2021fast,yang2022anarchic,koloskova2022sharper} and identical computation rate~\cite{Dmitrii21arbitrary}.
Moreover, several papers explore unique scenarios of client participation. 
For instance, \cite{chen2020optimal} selects the optimal client subset to minimize gradient estimation errors. 
The studies by \cite{malinovsky2023federated} and \cite{cho2023convergence} investigated cyclic client participation. Additionally, in \cite{wang2023lightweight}, optimal weights for each client are determined based on the estimated probabilities of their participation.
In contrast, this paper addresses a more practical worst-case scenario in FL -- ``incomplete client participation.'' 
This phenomenon may arise from various heterogeneous factors, as discussed in Section~\ref{sec: Introduction}.

\textbf{2) Domain Adaptation:}
Since incomplete client participation induces a gap between the dataset distribution used for FL training and the true data population distribution across all clients, our work is also related to the field of domain adaptation.
Domain adaptation focuses on the learnability of a model trained in one source domain but applied to a different and related target domain.
The basic approach is to quantify the error in terms of the source domain plus the distance between source and target domains.
Specifically, let $P$ and $Q$ be the target and source distributions, respectively.
Then, the generalization error is expressed as $\mc{O}(\mc{A}(n_Q)) + dist(P, Q)$, where $\mc{A}(n_Q)$ is an upper bound of the error dependent on the total number of samples in $Q$. 
Widely-used distance measures include $d_\mc{A}$-divergence ~\cite{ben2010theory,david2010impossibility} and $\mc{Y}$-discrepancy ~\cite{mansour2009domain,mohri2012new}.
We note, however, that results in domain adaptation is not directly applicable in FL with incomplete client participation, since doing so yields an overly pessimistic bound.
Specifically, the error based on domain adaptation remains non-zero for asymptotically small distance $dist(P, Q)$ between $P$ and $Q$ even with infinite many samples in $n_Q$ (i.e., $\mc{A}(n_Q) \rightarrow 0$).
In this paper, rather than directly using results from domain adaptation, we establish a much {\em sharper} upper bound (see Section~\ref{sec: PACLearning}).
A closely related work is \cite{hanneke2019value}, which proposed a new notion of discrepancy between source and target distributions.
However, this work considers {\em non-overlapping} support between $P$ and $Q$, while we focus on {\em overlapping} support naturally implied by FL (see Fig.~\ref{fig:FALvsDA} in Section~\ref{subsec:SA-FL}).


\section{PAC-Learnability of Federated Learning with Incomplete Client Participation} \label{sec: PACLearning}

In this section, we first focus on understanding the impacts of incomplete client participation on conventional FL 
in Section~\ref{subsec: FL}.
This will also pave the way for studying SA-FL later in Section~\ref{subsec:SA-FL}.
In what follows, we start with FL formulation and some definitions in statistical learning that are necessary to formulate and prove our main results.


The goal of an $M$-client FL system is to minimize the following loss function $F(\x) = \mb{E}_{i \sim \mc{P}} [F_i(\x)]$,
where $F_i(\x) \triangleq \mb{E}_{\xi ~\sim P_i} [f_i(\x, \xi)]$.
Here, $\mc{P}$ represents the distribution of the entire client population, $\x \in \mb{R}^d$ is the model parameter, $F_i(\x)$ represents the local loss function at client $i$, and $P_i$ is the underlying distribution of the local dataset at client $i$.
In general,  due to data heterogeneity, $P_i \neq P_j$ if $i \neq j$.
However, the loss function $F(\x)$ or full gradient $\nabla F(\x)$ can not be directly computed since the exact distribution of data is unknown in general.
Instead, one often considers the following empirical risk minimization (ERM) problem in the finite-sum form based on empirical risk $\hat{F}(\x)$: 
$$\min_{\x \in \mb{R}^d} \hat{F}(\x) \triangleq  \sum_{i \in [M]} \alpha_i \hat{F}_i(\x),  \hat{F}_i(\x) \triangleq \sum_{\xi \in S_i} f_i(\x, \xi),$$
where $S_i$ is a local dataset at client $i$ with cardinality $|S_i|$, whose samples are independently and identically sampled from distribution $P_i$, and $\alpha_i = |S_i|/(\sum_{j \in [M]}|S_j|)$ (hence $\sum_{i \in [M]} \alpha_i = 1$).
For simplicity, we consider the balanced dataset case: $\alpha_i = {1}/{M}, \forall i \in [M]$. 
Next, we state several definitions from statistical learning~\cite{mohri2018foundations}.
\begin{defn} [Generalization and Empirical Errors]
    Given a hypothesis $h \in \mc{H}$, a target concept $f$, an underlying distribution $\mc{D}$ and a dataset $S$ i.i.d. sampled from $\mc{D}$ ($S \sim \mc{D}$), the generalization error and empirical error of $h$ are defined as follows: $\mc{R}_{\mc{D}}(h, f) = \mb{E}_{(x, y) \sim \mc{D}} l(h(x), f(x))$ and $\hat{\mc{R}}_{D}(h, f) =(1/|S|) \sum_{i \in S} l(h(x_i), f(x_i))$,
where $l(\cdot)$ is a valid loss function. 
\end{defn}
\vspace{-0.1in}
For simplicity, we will use $\mc{R}_{\mc{D}}(h)$ and $\hat{\mc{R}}_{D}(h)$ for generalization and empirical errors and omit target concept $f$.

\begin{defn} [Optimal Hypothesis]
    We define $h_{\mc{D}}^{*} = \underset{h \in \mc{H}}{\mathrm{argmin}} \mc{R}_{\mc{D}}(h)$ and $\hat{h}_{\mc{D}}^{*} = \underset{h \in \mc{H}}{\mathrm{argmin}} \hat{\mc{R}}_{\mc{D}}(h)$.
\end{defn}

\begin{defn} [Excess Error]
    The excess error and excess empirical error are defined as $\varepsilon_{\mc{D}}(h) = \mc{R}_{\mc{D}}(h) - \mc{R}_{\mc{D}}(h_{\mc{D}}^{*})$, and $\hat{\varepsilon}_{\mc{D}}(h) = \hat{\mc{R}}_{\mc{D}}(h) - \hat{\mc{R}}_{\mc{D}}(\hat{h}_{\mc{D}}^{*})$, respectively.
\end{defn}

\subsection{Conventional Federated Learning with Incomplete Client Participation} \label{subsec: FL}
With the above notations, we now study conventional FL with incomplete client participation (Q1).
Consider an FL system with $M$ clients in total.
We let $P$ denote the underlying joint distribution of the entire system, which can be decomposed into the summation of the local distributions at each client, i.e., $P = \sum_{i \in [M]} \lambda_i P_i$, where $\lambda_i >0$ and $\sum_{i \in [M]} \lambda_i = 1$.
We assume that each client $i$ has $n$ training samples i.i.d. drawn from $P_i$, i.e., $|S_i| = n, \forall i \in [M]$.
Then, $S = \{(x_i, y_i), i \in [M \times n] \}$ can be viewed as the dataset i.i.d. sampled from the joint distribution $P$.
We consider an incomplete client participation setting, where $m \in [0, M)$ clients participate in the FL training as a result of some client sampling/participation process $\mc{F}$. We let $\mc{F}(S)$ represent the data ensemble actually used in training and $\mc{D}$ denote the underlying distribution corresponding to $\mc{F}(S)$. 
For convenience, we define the notion $\omega  = \frac{m}{M}$ as the {\em FL system capacity} (i.e., only $m$ clients participate in the training).
For FL with incomplete client participation, we establish the following fundamental performance limit of any FL learner in general.
For simplicity, we use binary classification with zero-one loss here, but it is already sufficient to establish the PAC learnability lower limit.

\begin{defn}
A concept class $\mathcal{C}$ is said to be PAC learnable if there exists an algorithm $\mathcal{A}$ and a polynomial function $poly(\cdot, \cdot, \cdot, \cdot)$ such that for any $\epsilon > 0$ and $\delta > 0$, and for any distribution $\mathcal{D}$ over the instance space $\mathcal{X}$, the algorithm, with probability at least $1 - \delta$, outputs a hypothesis $h$ such that: $\mathbb{P}_{S \sim D}(R(h_S) \leq \epsilon) \geq 1 - \delta,$ where $R(h)$ denotes the generalization error of hypothesis $h$ returned by the algorithm. When such an algorithm $\mathcal{A}$ exists, it is called a PAC-learning algorithm for $\mathcal{C}$.
\end{defn}

In plain language, being PAC-learnable requires the hypothesis (or the model) returned by the algorithm after observing enough number of data samples is approximately correct (error at most $\epsilon$) with high probability (at least $1 - \delta$).

\begin{restatable} [Impossibility Theorem] {theorem} {FLLB}
\label{thm:LB}
Let $\mc{H}$ be a non-trivial hypothesis space and $\mc{L}: (\mc{X}, \mc{Y})^{(m \times n)} \rightarrow \mc{H}$ be the learner for an FL system.
There exists a client participation process $\mc{F}$ with FL system capacity $\omega$, a distribution $P$, and a target concept $f \in \mc{H}$ with $\min_{h \in \mc{H}} \mc{R}_{P} (h, f) = 0$, such that
$\mb{P}_{S \sim P} \big[\mc{R}_P(\mc{L}(\mc{F}(S), f)) > \frac{1 - \omega }{8} \big] > \frac{1}{20}$.
\end{restatable}

\vspace{-.2in}
\begin{proof}[Proof Sketch]
The proof is based on the method of induced distributions in~\cite{bshouty2002pac,mohri2018foundations,konstantinov2020sample}.
We first show that the learnability of an FL system is equivalent to that of a system that arbitrarily selects $mn$ out of $Mn$ samples in the centralized learning.
Then, for any learning algorithm, there exists a distribution $P$ such that dataset $\mc{F}(S)$ resulting from incomplete participation and seen by the algorithm is always distributed identically for any target functions.
Due to space limitation, we relegate the full proof to appendix.
\end{proof}

\vspace{-0.1in}
Given the system capacity $\omega \in (0,1)$, the above theorem characterizes the worst-case scenario for FL with incomplete client participation.
It says that for any learner (i.e., algorithm) $\mc{L}$, there exist a bad client participation process $\mc{F}$ and distributions $P_i, i \in [M]$ over target function $f$, for which the error of the hypotheses returned by $\mc{L}$ is constant with non-zero probability.
In other words, FL with incomplete client participation is {\em not PAC-learnable}.
One interesting observation here is that the lower bound is {\em independent} of the number of samples per client $n$.
This indicates that even if each client has {\em infinitely many} samples ($n \rightarrow \infty$), it is impossible to have a zero-generation-error learner under the incomplete client participation (i.e., $\omega \in (0,1)$).
Note that this fundamental result relies on two conditions: {\em heterogeneous} dataset and {\em arbitrary} client participation.
Under these two conditions, there exists a worst-case scenario where the underlying distribution $\mc{D}$ of the participating data $S_{\mc{D}} = \mc{F}(S)$ deviates from the ground truth $P$, thus yielding a non-vanishing error.

This result sheds light on system and algorithm design for FL. That is, how to motivate client participation in FL effectively and efficiently: the participating client's data should be comprehensive enough to model the complexity of the joint distribution $P$ to close the gap between $\mc{D}$ and $P$.
Note that this result is not contradictory to previous works where the convergence of FedAvg is guaranteed, since this theorem is not applicable for homogeneous (i.i.d.) datasets or uniformly random client participation.
As mentioned earlier, most of the existing works rely on at least one of these two assumptions.
However, none of these two assumptions hold for conventional FL with incomplete client participation in practice.
In addition to system heterogeneity, other factors such as Byzantine attackers could also render incomplete client participation.
For example, even for full client participation in FL, if part of the clients are Byzantine attackers, the impossibility theorem also applies.
Thus, our impossibility theorem also justifies the empirical use of server-assisted federated learning (i.e., FL with server-side auxiliary data) to build trust~\cite{Cao2021FLTrustBF}.

\subsection{The PAC-Learnability of Server-Assisted Federated Learning (SA-FL)} \label{subsec:SA-FL}

The intuition of SA-FL is to utilize a dataset $T$ i.i.d. sampled from distribution $P$ with cardinality $|T| = n_T$ as a vehicle to correct potential distribution deviations due to incomplete client participation.
By doing so, the server steers the learning by a small number of representative data, while the clients assist the learning by federation to leverage the huge amount of privately decentralized data ($n_S \gg n_T$).
%

For SA-FL, we consider the same incomplete client participation setting that induces a dataset $S_{\mc{D}} \sim \mc{D}$ with cardinality $n_S$ and $\mc{D} \neq P$ (i.e., Q2).
As a result, the learning process is to minimize $\mc{R}_{P}(h)$ by utilizing $(\mathcal{X}, \mathcal{Y})^{n_T + n_S}$ to learn a hypothesis $h \in \mc{H}$.
For notional clarity, we assume the joint dataset $S_{Q} = (S_{\mc{D}} \cup T) \sim Q$ with cardinality $n_T + n_S$ for some distribution $Q$.
Before deriving the generalization error bound for SA-FL, we state the following assumptions and definitions. 

\begin{assum} [Noise Condition] \label{assum: NC}
    Suppose $h_P^{*}$ and $h_Q^{*}$ exist. 
    There exist $\beta_P, \beta_Q \in [0, 1]$  and $\alpha_P, \alpha_Q > 0$ s.t.,
        $\mb{P}_{x \sim P} (h(x) \neq h_P^{*}(x)) \leq \alpha_P [\varepsilon_{P}(h)]^{\beta_P}$, 
        $\mb{P}_{x \sim Q} (h(x) \neq h_Q^{*}(x)) \leq \alpha_q [\varepsilon_{Q}(h)]^{\beta_Q}$.
\end{assum}
This assumption is a traditional noise model known as the Bernstein class condition, which has been widely used in the literature~\cite{massart2006risk,koltchinskii2006local,hanneke2016refined}.

\begin{assum}[$(\alpha,\beta)$-Positively-Related] \label{assum: PR}
    Distributions $P$ and $Q$ are said to be $(\alpha, \beta)$-positively-related if there exist constants $\alpha \geq 0$ and $\beta \geq 0$ such that
        $| \varepsilon_{P}(h) - \varepsilon_{Q}(h) | \leq \alpha [\varepsilon_{Q}(h)]^{\beta}, \forall h \in \mc{H}$.
\end{assum}

Assumption~\ref{assum: PR} specifies a stronger constraint between distributions $P$ and $Q$.
It implies that the difference of excess error for one hypothesis $h \in \mc{H}$ between $P$ and $Q$ is bounded by the excess error of $Q$ in some exponential form.
Assumption~\ref{assum: PR} is one of the major {\em novelty} in our paper and unseen in the literature.
We note that this $(\alpha,\beta)$-positively-related condition is a mild condition. 
To see this, consider the following ``one-dimensional'' example for simplicity. Let $\mathcal{H}$ be the class of hypotheses defined on the real line: $\{ h_t = t, t \in R \}$, and let two uniform distributions be $P := \mathcal{U}[a, b]$ and $Q := \mathcal{U}[a', b']$. Due to the incomplete client sampling in FL, the support of $Q$ is a subset of that of $P$, i.e., $a \leq a' \leq b' \leq b$. Denote the target hypothesis $t^* \in [a', b']$. Then, for any hypothesis $h_t$ with threshold $t$, we have $\epsilon_P(h_t) = \frac{| t - t^* |}{b - a}$ and $\epsilon_Q(h_t) = \frac{| t - t^* |}{b' - a'}$. That is, our "$(\alpha,\beta)$-Positively-Related" holds for $\alpha = 1 - \frac{b' - a'}{b - a}$ and $\beta = 1$. 
The above ``one-dimensional'' example can be further extended to general high-dimensional cases as follows. 
Intuitively, the difference of excess errors of $P$ and $Q$ (i.e., $| \epsilon_P(h) -  \epsilon_Q(h) |$) is a function in the form of $\int_S |Q_X - P_X|dS$ for a common support domain $S \subset supp(Q)$. Thus, the ``$(\alpha,\beta)$-Positively-Related'' condition can be written as $| \int_S Q_X dS - \int_S P_X dS | \leq \alpha (\int_S Q_X)^{\beta}$. If distribution $Q$ has more probability mass over $S$ than distribution $P$, choosing $\beta = 1$ and $\alpha$ to be a sufficiently large constant clearly satisfies the $(\alpha,\beta)$-positively-related condition. Otherwise, letting $\beta \rightarrow 0$ and choosing $\alpha$ to be a sufficiently large constant satisfies the $(\alpha,\beta)$-positively-related condition with probability one.

With the above assumption and definition, we have the following generation error bound for SA-FL, which shows that SA-FL is PAC-learnable:

\begin{restatable} [Generalization Error Bound for SA-FL] {theorem} {FALbound}
    \label{thm:FAL_GB}
        For an SA-FL system with arbitrary system and data heterogeneity,
        if distributions $P$ and $Q$ satisfy Assumption~\ref{assum: NC} and \ref{assum: PR}, then with probability at least $1-\delta$ for any $\delta \in (0,1)$, it holds that
        \begin{eqnarray} \label{eqn:FAL_GB}
            \varepsilon_{P}(\hat{h}_Q^*) \!\!=\! \widetilde{\mc{O}} \left( \left(\frac{d_{\mc{H}}}{n_T+n_S} \right)^{\frac{1}{2-\beta_Q}} \!\!\!\!\!\!\!+\!\!  \left(\frac{d_{\mc{H}}}{n_T+n_S} \right)^{\frac{\beta}{2-\beta_Q}} \right),
        \end{eqnarray}
        where $d_{\mc{H}}$ is the finite VC dimension for hypotheses class $\mc{H}$.
\end{restatable}

Note the generalization error bound of centralized learning is $\widetilde{\mc{O}} ( (\frac{d_{\mc{H}}}{n})^{\frac{1}{2-\beta_Q}})$ (hiding logarithmic factors) with $n$ samples in total and noise parameter $\beta_Q$~\cite{hanneke2016refined}.
Note that when $\beta \geq 1$, the first term in Eq.~(\ref{eqn:FAL_GB}) dominates.
Hence, Theorem~\ref{thm:FAL_GB} implies that the generalization error bound in this case for SA-FL {\em matches} that of centralized learning (with dataset size $n_T + n_S$).
Meanwhile, for $0 < \beta < 1$, compared with solely training on server's dataset $T$, SA-FL exhibits an improvement from $\widetilde{\mc{O}} ( (\frac{1}{n_T})^{\frac{1}{2-\beta_Q}})$ to $\widetilde{\mc{O}} ( (\frac{1}{n_T + n_S})^{\frac{\beta}{2-\beta_Q}})$.

Note that SA-FL shares some similarity with the domain adaptation problem, where the learning is on $Q$ but the results will be adapted to $P$.
In what follows, we offer some deeper insights between the two by answering two key questions: {\em 1) What is the difference between SA-FL and domain adaptation (or transfer learning)?} and {\em 2) Why is SA-FL from $Q$ to $P$ PAC-learnable, but FL  from $D$ to $P$ with incomplete client participation not PAC-learnable (as indicated in Theorem~\ref{thm:LB})?}

\begin{figure} 
    \vspace{-0.05in}
    \centering
    \includegraphics[width=0.4\textwidth]{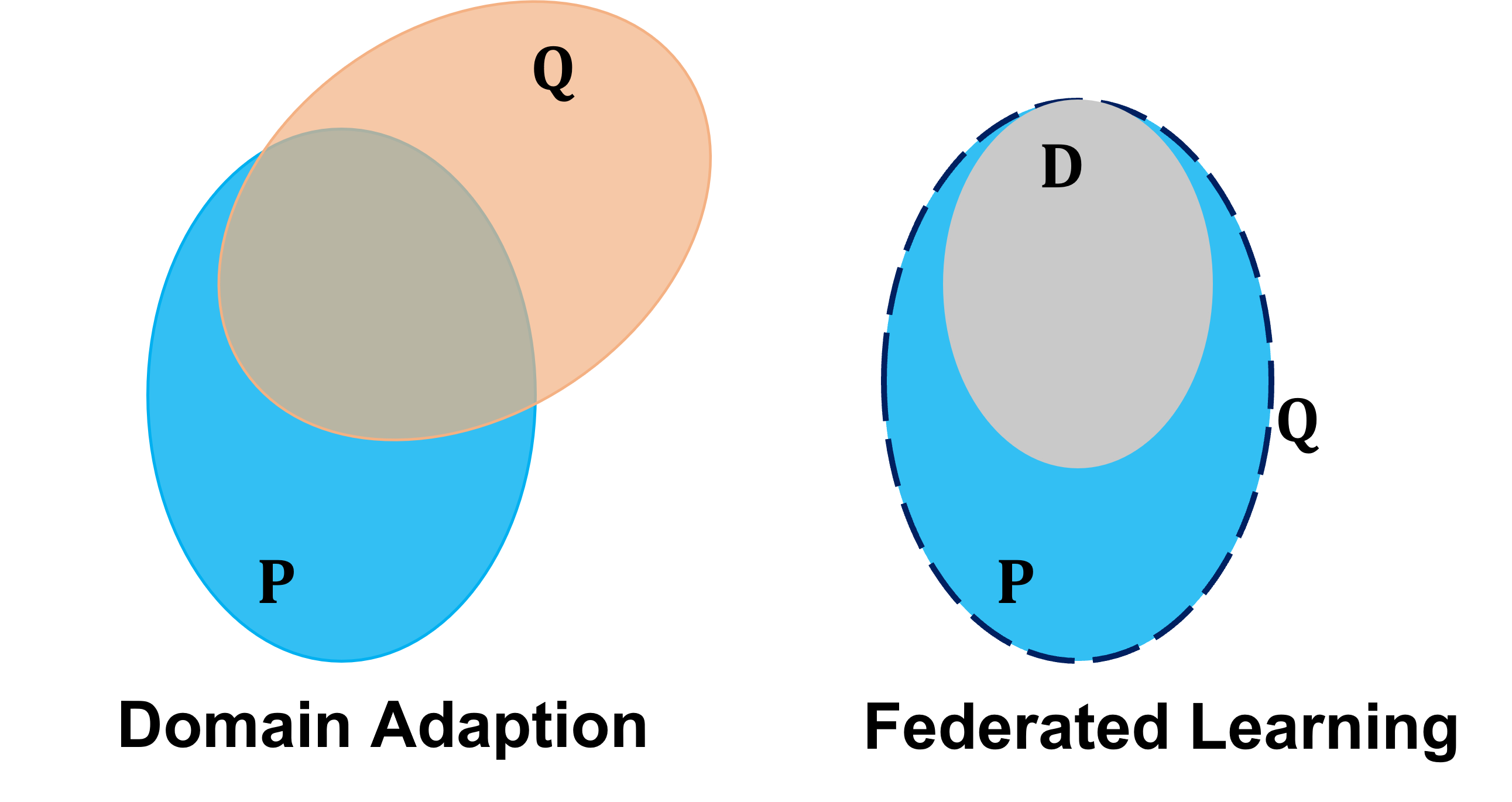}
    \vspace{-0.15in}
    \caption{Diagram of distribution supports for domain adaptation and federated learning.}
    \label{fig:FALvsDA}
    \vspace{-0.7cm}
\end{figure}

To answer these questions, we illustrate the distribution relationships for domain adaptation and federated learning, in Fig.~\ref{fig:FALvsDA}, respectively.
In domain adaptation, the target $P$ and source $Q$ distributions often have overlapping support but there also exists {\em distinguishable difference}.
In contrast, the two distributions $P$ and $Q$ in SA-FL happen to share exactly the {\em same support} with different density, since $Q$ is a {\em mixture} of $D$ and $P$.
As a result, the known bounds in domain adaptation (or transfer learning) are pessimistic for SA-FL. 
For example, the $dist(P, Q)$ in $d_{\mc{A}}$-divergence and $\mc{Y}$-divergence both have non-negligible gaps when applied to SA-FL.
Here in Theorem~\ref{thm:FAL_GB}, we provide a generalization error bound in terms of the total sample size $n_T + n_S$, thus showing the benefit of SA-FL.

Moreover, for SA-FL, only the auxiliary dataset $T \overset{i.i.d.}{\sim} P$ is directly available to the server.
The clients' datasets could be used in SA-FL training, but they are not directly accessible due to privacy constraints. 
Thus, previous methods in domain adaptation (e.g., importance weights-based methods in covariate shift adaptation~\cite{sugiyama2007covariate,sugiyama2007direct}) are {\em not} applicable since they require the knowledge of density ratio between training and test datasets.

The key difference between FL and SA-FL lies in relations among $D, P$ and $Q$.
For FL, the distance between $D$ and $P$ with incomplete participation could be large due to system and data heterogeneity in the worst-case.
More specifically, the support of $D$ could be narrow enough to miss some part of $P$, resulting in non-vanishing error as indicated in Theorem~\ref{thm:LB}.
For SA-FL, distribution $Q$ is a mixture of $P$ and $D$ ($Q = \lambda_1 D + \lambda_2 P$, with $\lambda_1,\lambda_2 \geq 0$, $\lambda_1 + \lambda_2 = 1$), thus having the same support with $P$.
Hence, under Assumption~\ref{assum: PR}, the PAC-learnability is guaranteed.
Although we provide a promising bound to show the PAC-learnability of SA-FL in Theorem~\ref{thm:FAL_GB}, the superiority of SA-FL over training solely with dataset $T$ in server (i.e., $\widetilde{\mc{O}} ((\frac{1}{n_T} )^{\frac{1}{2-\beta_P}})$) is not always guaranteed as $\beta \rightarrow 0$ (i.e., $Q$ becomes increasingly different from $P$).
In what follows, we reveal under what conditions could SA-FL perform {\em no worse than} centralized learning.

\begin{restatable} [Conditions of SA-FL Being No Worse Than Centralized Learning] {theorem} {FALCL}
    \label{thm:FAL_CL_UB}
        Consider an SA-FL system with arbitrary system and data heterogeneity.
        If Assumption~\ref{assum: NC} holds and additionally $\mc{\hat{R}}_P(\hat{h}_Q^*) \leq \mc{\hat{R}}_P(h_Q^*)$ and $\varepsilon_{P}(h_Q^*) = \mc{O}(\mathcal{A}(n_T, \delta))$, where $\mathcal{A}(n_T, \delta)=\frac{d_{\mc{H}}}{n_T} \log(\frac{n_T}{d_{\mc{H}}} + \frac{1}{n_T} \log(\frac{1}{\delta}))$, then with probability at least $1-\delta$ for any $\delta \in (0,1)$, it holds that
$\varepsilon_{P}(\hat{h}_Q^*) = \widetilde{\mc{O}} \left( \left(d_{\mc{H}}/n_T \right)^{\frac{1}{2-\beta_P}} \right)$. 
\end{restatable}

Here, we remark that $\varepsilon_{P}(h_Q^*) = \mc{O}(\mathcal{A}(n_T, \delta))$ is a weaker condition than the 
$\varepsilon_{P}(h_Q^*) = 0$ condition and the covariate shift assumption ($P_{Y | X} = Q_{Y | X}$) used in the transfer learning literatures~\cite{hanneke2019value,hanneke2020no}.
Together with the condition $\mc{\hat{R}}_P(\hat{h}_Q^*) \leq \mc{\hat{R}}_P(h_Q^*)$, the following intermediate result holds:
$\mc{\hat{R}}_P(\hat{h}_Q^*) - \mc{\hat{R}}_P(h_P^*) = \mc{O}(A(n_T, \delta))$ (see Lemma~\ref{lm:Lemma_appx} in the supplementary material).
Intuitively, this states that ``if $P$ and $Q$ share enough similarity, then the difference of excess empirical error between $\hat{h}_Q^*$ and $h_P^*$ on $P$ can be bounded.'' 
Thus, the excess error of $\hat{h}_Q^*$ shares the same upper bound as that of $\hat{h}_P^*$ in centralized learning.
Therefore, Theorem~\ref{thm:FAL_CL_UB} implies that, under mild conditions, SA-FL guarantees the same generalization error upper bound as that of centralized learning, hence being ``no worse than'' centralized learning with dataset $T$.

\begin{rem}{\em
Note that the assumption of having a server-side dataset is not restrictive due to the following reasons. First, such datasets are already available in many FL systems: although not always necessary for
training, an auxiliary dataset is often needed for defining FL tasks (e.g., simulation prototyping) before training and model checking after
training (e.g., quality evaluation and sanity checking)~\cite{mcmahan2021fl,wang2021field}. Also, obtaining an auxiliary dataset is affordable since the number of data points required is relatively small, and hence the cost is low. Then, SA-FL can be easily achieved or even with manually labelled data thanks to its small size.
It is also worth noting that many works have used such auxiliary datasets in FL for security~\cite{Cao2021FLTrustBF},
incentive design ~\cite{wang2019measure}, and knowledge distillation ~\cite{cho2021personalized}.
}
\end{rem} 

\begin{rem}{\em 
It is also worth pointing out that, for ease of illustration, Theorem~\ref{thm:FAL_GB}--\ref{thm:FAL_CL_UB} are based on the assumption that the auxiliary dataset $T \overset{i.i.d.}{\sim} P$. 
Nonetheless, it is of practical importance to consider the scenario where $T$ is sampled from a related but slightly different distribution $P'$ rather than the target distribution $P$ itself. 
In fact, the above assumption could be relaxed to $T \overset{i.i.d.}{\sim} P^{'}$ for any $P'$ as long as the mixture distribution $Q = \lambda_1 D + \lambda_2 P'$ is $(\alpha,\beta)$-positively-related with $P$. 
Under such condition, we can show that the main results in Theorem~\ref{thm:FAL_GB}--\ref{thm:FAL_CL_UB} still hold. 
}
\end{rem}


\section{The \alg Algorithm for SA-FL} \label{sec: Algorithm}

\begin{algorithm}[t!]
    \caption{The \alg Algorithm for SA-FL.} \label{alg:FAA} 
    \begin{algorithmic}[1]
    \STATE{Initialize model $\x_0$, iteration index $t=0$.}
    \FOR{$r = 0, \cdots, R - 1$}
        \STATE With probability $q$: \hfill $\bigstar $  client update round $r \in \mathcal{T}_c$ 
            \STATE \hspace{1em} The server samples clients $S_r$. 
            \STATE \hspace{1em} Each client $i \in S_r$ computes in parallel:
            \STATE \hspace{2em} $\x_{r, k+1}^i = \x_{r, k}^i - \eta_c \nabla F_i(\x_{r, k}^i, \xi_{r, k}^i), k \in [K]$, 
            \STATE \hspace{2em} starting from $\x_{r, 0}^i = \x_{r}$.
            \STATE \hspace{2em} Send $\x_{r}^i = \x_{r, K+1}^i$ to server.
            \STATE \hspace{1em} Server updates: $\x_{r+1} = \frac{1}{| S_r |} \sum_{i \in S_r} \x_{r}^i$.
        \STATE Otherwise \hfill $\bigstar $ server update round $r \in \mathcal{T}_s$
            \STATE \hspace{1em} Server updates: $\x_{r+1} = \x_r - \eta_s \nabla F(\x_r, \xi_r)$. 
    \ENDFOR
    \end{algorithmic}
    \end{algorithm}

In Section~\ref{sec: PACLearning}, we have shown that SA-FL is PAC-learnable with incomplete client participation.
In this section, we turn our attention to the {\em training} of the SA-FL regime with incomplete client participation (i.e., Q3), which is also under-explored in the literature.
First, we note that the standard FedAvg algorithm may fail to converge to a stationary point with incomplete client participation as indicated by previous works~\cite{yang2022anarchic}.
Now with SA-FL, we aim to answer the following questions:

\vspace{-.15in}
{\em 
\begin{list}{\labelitemi}{\leftmargin=1.0em \itemindent=-0.1em \itemsep=.0em}
\item[1)] Under SA-FL, how should we appropriately use the server-side dataset to develop training algorithms in the SA-FL regime with provable convergence guarantees?

\item[2)] If Question 1) can be resolved, could we further achieve the same convergence rate in SA-FL training with incomplete client participation as that in traditional FL with ideal client participation?

\end{list}
}

\vspace{-.1in}
In this section, we resolve the above questions affirmatively by proposing a new algorithm called \alg (\ul{s}erver-\ul{a}ssisted \ul{f}ederated \ul{a}ve\ul{r}ag\ul{i}ng) for SA-FL with theoretically provable convergence guarantees.
As shown in Algorithm~\ref{alg:FAA}, \alg contains two options in each round, client update option or global server update option.
For a communication round $r \in \{ 0, \cdots, R - 1 \}$, with probability $q \in [0,1]$, the client update option is chosen (i.e., $r \in \mathcal{T}_c$), where local updates are executed by clients in the current participating client set $S_r$ in a similar fashion as the FedAvg~\cite{mcmahan2017communication}.
Specifically, the client update option performs the following three steps:
1) Server samples a subset of clients $S_r$ as in conventional FL and synchronizes the latest global model $\x_r$ with each participating clients in $S_r$ (Line~4);
2) All participating clients initialize their local models as $\x_r$ and then perform $K$ local steps following the stochastic gradient descent (SGD) method. 
Then, each participating client sends its locally updated model $\x_{r}^i = \x_{r, K+1}^i$ back to the server (Lines~5-8);
3) Upon receiving the local update $\x_{r}^i$, the server aggregates and updates the global model (Line~9).
On the other hand, with probability $1-q$, the server update option is chosen (i.e,. $r \in \mathcal{T}_s$), where the server updates the global model with its auxiliary data following the SGD (Line~11).

We note that \alg can be viewed as a mixture of the FedAvg algorithm with client-side datasets (cf. the client update option) and a centralized SGD algorithm using the server-side dataset only (cf. the server update option), which are governed by a probability parameter $q$.
The basic idea of this two-option approach is to leverage client-side parallel computing to accelerate the training process, while using the server-side dataset to mitigate the bias caused by incomplete client participation. 
We will show later that, by appropriately choosing the $q$-value, \alg simultaneously achieves the stationary point convergence and linear convergence speedup. 
Before presenting the convergence performance results, we first state three commonly used assumptions in FL.

\begin{assum}($L$-Lipschitz Continuous Gradient) \label{assum: smooth}
	There exists a constant $L > 0$, such that $ \| \nabla F_i(\x) - \nabla F_i(\y) \| \leq L \| \x - \y \|$, $\forall i \in [M], \x, \y \in \mathbb{R}^d$.
\end{assum}

\begin{assum}
(Unbiased Stochastic Gradients with Bounded Variance)
\label{assum: unbias}
	The stochastic gradient calculated by the client or server is unbiased with bounded variance: for server, $\mathbb{E} [\nabla F(\x, \xi)] = \nabla F(\x)$ and $\mathbb{E} [\| \nabla F(\x, \xi)- \nabla F(\x) \|^2] \leq \sigma_s^2$; for each client $i \in [M]$, $\mathbb{E} [\nabla F_i(\x, \xi)] = \nabla F_i(\x)$, and $\mathbb{E} [\| \nabla F_i(\x, \xi)- \nabla F_i(\x) \|^2] \leq \sigma^2$.
\end{assum}

\begin{assum}(Bounded Gradient Dissimilarity) \label{assum: gradient}
	$\| \nabla F_i(\x)- \nabla F(\x) \|^2 \leq \sigma_G^2, \forall i \in [M]$.
\end{assum}

With the assumptions above, we state the main convergence result of \alg for non-convex functions as follows:
\begin{restatable} [Convergence Rate for \alg in Non-Convex Functions] {theorem} {FAA} \label{thm:convergence_FAA}
    Under Assumptions~\ref{assum: smooth} - ~\ref{assum: gradient}, if $\eta_c \leq \frac{1}{4\sqrt{30} LK}$, $\eta_c = \frac{2 \eta_s}{K}$, and $q \leq 1 / \left( \frac{4 \sigma_G^2 - 4G_2 (\frac{1}{2K^2} - \frac{2L \eta_s^2}{K^2})}{(1 - L \eta_s) G_1} + 1 \right)$, 
    then, the sequence $\{ \x_r \}$ generated by \alg satisfies:
    \begin{align*}
        \frac{1}{R} \sum_{r = 1}^R \mb{E} \| \nabla F(\x_r) \|^2 \leq \frac{2 (F(\x_0) - F(\x^*))}{R \eta_s} + \delta, 
    \end{align*}

    where $\delta = L \eta_s (1-q) \sigma_s^2 + \frac{80 q L^2 \eta_s^2}{K }(\sigma^2 + 6K \sigma_G^2) + \frac{8L q \eta_s}{mK} \sigma^2$,  
    $G_1 = \max_{r \in \mathcal{T}_s} \| \nabla F(\x_r) \|^2$, and $G_2 = \max_{r \in \mathcal{T}_c} \left\| \frac{1}{m} \sum_{i \in [m]} \sum_{k \in [K]} \nabla F_i(\x_{r, k}^i) \right\|^2$.
\end{restatable}

\begin{rem}{\em
Theorem~\ref{thm:convergence_FAA} says that, by using the server-side update with an appropriately chosen $q$-value, \alg effectively mitigates the bias that arises from incomplete client participation. 
With proper probability $q$,
\alg guarantees stationary point convergence in non-convex functions.
%
}
\end{rem}

By choosing parameters $q$ and the learning rate $\eta$ appropriately, Theorem~\ref{thm:convergence_FAA} immediately implies the following rate:
\begin{restatable} {corollary} {FAA_order} \label{cor:convergence_FAA}
    If $\eta_s = \frac{1}{\sqrt{R}}$,
    \alg achieves an $\mathcal{O} ( \frac{1}{\sqrt{R}} )$ convergence rate to a stationary point. 
    If we can further assume the stochastic gradient noise at server's side $\sigma_s^2 = \mathcal{O}(\frac{1}{mK} \sigma^2)$ or $q = \Omega (1 - \frac{1}{mK})$,
    \alg achieves an $\mathcal{O} ( \frac{1}{\sqrt{mKR}} )$ convergence rate to a stationary point, implying a linear speedup of convergence in terms of $m$ and $K$. 
\end{restatable}


For strongly convex functions, we have the following convergence results for \alg:
\begin{restatable} [Convergence Rate for \alg in Strongly Convex Functions] {theorem} {FAS} \label{thm:convergence_FAA_2}
    Under Assumptions~\ref{assum: smooth} - ~\ref{assum: gradient} and assume each function $F_i$ is $\mu-$strongly convex, if $\eta_s \leq \frac{2}{L + \mu}$ and $q \leq 1 / \left( 1 + \frac{\frac{4\bar{\eta}}{\mu^2} \left(1 + \frac{30 L \bar{\eta}}{\mu}(1 + \frac{2L \bar{\eta}}{\mu}) \right) G_3 - \frac{4}{ \mu} G_4}{\left(\frac{1}{L + \mu} - \frac{(L+\mu)^2 \bar{\eta}}{4L^2 \mu^2}\right) G_3 } \right)$,
    then, the sequence $\{ \x_r \}$ generated by \alg satisfies:
    \begin{align*}
        \mathbb{E} \| \x_{R} - \x_* \|^2 
        &\leq (1 - \bar{\eta})^R \| \x_0 - \x_* \|^2 +  \bar{\delta},
    \end{align*}
    where 
    $\bar{\delta} \!=\! \frac{8 q \eta_c K}{\mu} \sigma_G^2  \!+\! \frac{2 q \eta_c}{\mu m} \sigma^2 + \frac{4 q  L(1 + K \eta_c L)}{\mu} \times \left[5K \eta_c^2 (\sigma^2 + 6K \sigma_G^2) \right] + (1 - q) \frac{L + \mu}{2 L \mu} \eta_s \sigma_s^2$, 
    $G_3 = \left\| \nabla F(\x_r) \right\|^2 $, $G_4 = \frac{1}{m} \sum_{i \in S_r} \left[ F_i(\x_r) - F_i(\x_*) \right]$, and $\bar{\eta} = \frac{\eta_c K \mu}{2} = \frac{2 \eta_s L\mu}{L + \mu}$.
\end{restatable}

The following results immediately follow from Theorem~\ref{thm:convergence_FAA_2}:
\begin{restatable} {corollary} {FAS_order} \label{cor:convergence_FAS}
    If $\eta_c = \Omega(\frac{log(R)}{R})$ and $\eta_s = \Omega(\frac{log(R)}{R})$, \alg achieves an $\tilde{\mathcal{O}}(\frac{1}{R})$ convergence rate. 
\end{restatable}

\begin{rem} {\em
In the strongly convex setting, Theorem~\ref{thm:convergence_FAA_2} shows that, with proper hyperparameters, \alg achieves convergence guarantees and can effectively mitigate the impacts of incomplete client participation. 
}
\end{rem}

\begin{rem}{\em
We note that \alg is a unifying framework that includes two classic algorithms as special cases under two extreme settings: i) the i.i.d. client-side data case and ii) the heterogeneous client-side data case with unbounded gradient dissimilarity. 
    In the i.i.d. case, the client-side data are homogeneous, i.e., $F_i(\x) = F$ and $\sigma_G = 0$. 
    In this ideal setting, we can simply choose $q=1$ and \alg reduces to the classical FedAvg algorithm. 
    In the heterogeneous case with unbounded gradient dissimilarity (i.e., $\sigma_G \rightarrow \infty$), 
    we can set $q = 0$ (i.e., $| \mathcal{T}_c | = 0$) such that \alg reduces to the centralized SGD algorithm.
    In this setting, Theorem~\ref{thm:convergence_FAA} and \ref{thm:convergence_FAA_2} recover the classic SGD bounds by cancelling the $\sigma_G$-dependent terms in the bound. 
}
\end{rem}

\begin{rem}{\em
Corollary~\ref{cor:convergence_FAA} and \ref{cor:convergence_FAS} suggest that, thanks to the two ``control knobs'', learning rate $\eta$ and $q$ in \alg, under mild conditions, we can avoid the limitation of conventional FL algorithms.
For example, FedAvg with incomplete client participation can only converge to an error ball dependent on the data heterogeneity parameter $\sigma_G$~\cite{yang2022anarchic}. 
In SA-FL, \alg with incomplete client participation can still achieve the same convergence rates as these of classic FedAvg algorithms with ideal client participation: $\mathcal{O} (1/\sqrt{mKR} )$ for non-convex functions~\cite{yang2021linearspeedup} and $\tilde{\mathcal{O}}(1/R)$ for strongly convex functions~\cite{Li2020convergence}. 
}
\end{rem}


\section{Numerical results} \label{sec: NumericalResults}

In this section, we conduct numerical experiments to verify our theoretical results using 1) logistic regression (LR) on MNIST dataset~\citep{lecun1998gradient}, 2) convolutional neural network (CNN) on CIFAR-10 dataset~\citep{krizhevsky2009learning}.
To simulate data heterogeneity, we distribute the data into each client evenly in a label-based partition, following the same process as in previous works~\citep{mcmahan2017communication, yang2021linearspeedup, Li2020convergence}.
As a result, we can use a parameter $p \in \{ 1, 2, 5, 10 \}$ to represent the classes of labels in each client’s dataset, which serves as an index of data heterogeneity level (non-i.i.d. index).
The smaller $p$-value, the more heterogeneous the data among clients.
To mimic incomplete client participation, we force $s$ clients to be excluded.
We can use $s \in \{0, 2, 4 \}$ as an index to represent the degree of incomplete client participation.
In our experiments, there are $M=10$ clients in total, and $m=5$ clients participate in the training in each communication round, who are uniformly sampled from the $M-s$ clients.
We use FedAvg without any server-side dataset as the baseline to compare with \alg with auxiliary data size $\{50, 100, 500, 1000 \}$ for MNIST and $\{ 500, 5000, 10000 \}$ for CIFAR10.
So in each dataset, we have at least $4 \times 3 \times 4 = 48$ sets of experiment for ablation study.
Due to space limitation, we highlight the key observations in this section, and relegate all other experimental details and results to the supplementary material.



\begin{table}[t]
\centering
\caption{Test accuracy (\%) for FedAvg.}
\label{table:FEDAVG_P}
\begin{small}
\begin{sc}
\begin{tabular}{cccccc}
\hline
\multirow{2}{*}{Dataset}  & \multirow{2}{*}{$s$} & \multicolumn{4}{c}{non-i.i.d. index ($p$)} \\ \cline{3-6} 
                          &                      & 10        & 5        & 2        & 1        \\ \hline
\multirow{3}{*}{MNIST}    & 0                    & 92.69     & 89.49    & 86.17    & 84.49    \\
                          & 2                    & 92.64     & 89.11    & 86.54    & 71.58    \\
                          & 4                    & 92.62     & 88.81    & 77.81    & 57.05    \\ \hline
\multirow{3}{*}{CIFAR-10} & 0                    & 81.12     & 79.42    & 78.22    & 75.7     \\
                          & 2                    & 79.97     & 78.54    & 76.68    & 64.56    \\
                          & 4                    & 77.78     & 75.55    & 67.34    & 50.7     \\ \hline
\end{tabular}
\end{sc}
\end{small}
\vspace{-0.15in}
\end{table}

\textbf{1) Performance Degradation of Incomplete Client Participation:}
As shown in Table~\ref{table:FEDAVG_P}, a distinct and non-trivial decline in performance is observed for FedAvg when confronted with incomplete client participation. 
As the value of $s$ increases, it signifies progressively more incomplete client participation.
Upon comparing scenarios with $s=0$ and $s=4$ , a significant reduction in test accuracy becomes apparent, reaching up to $27.44\%$ for MNIST and $25\%$ for CIFAR10.
It is noteworthy that such performance degradation is also contingent on data heterogeneity, denoted by $p$. In the case of IID data ($p=10$), only a negligible decrease in test accuracy is observed. For instance, in MNIST, the accuracy slightly drops from $92.69\%$ for $s=0$ to $92.62\%$ for $s=4$. 
In CIFAR10, the accuracy decreases from $81.12\%$ for $s=0$ to $77.78\%$ for $s=4$.
These results empirically validate the worst-case analysis in Theorem~\ref{thm:LB} and serve as the primary motivation for the development of SA-FL.

\begin{table}[t]
\centering
\caption{Test accuracy improvement (\%) for \alg ($q=0.8$) compared with FedAvg. `-' means ``no statistical difference within $2\%$ error bar''.}
\label{table:SAFL_vs_FedAvg}
\begin{small}
\begin{sc}
\begin{tabular}{ccllll}
\hline
\multirow{2}{*}{Dataset}  & \multirow{2}{*}{$s$} & \multicolumn{4}{c}{non-i.i.d. index ($p$)}                                                     \\ \cline{3-6} 
                          &                      & \multicolumn{1}{c}{10} & \multicolumn{1}{c}{5} & \multicolumn{1}{c}{2} & \multicolumn{1}{c}{1} \\ \hline
\multirow{3}{*}{MNIST}    & 0                    & -                      & -                     & 3.13                  & 4.47                  \\
                          & 2                    & -                      & -                     & 2.01                  & 16.53                 \\
                          & 4                    & -                      & -                     & 10.69                 & 31.07                 \\ \hline
\multirow{3}{*}{CIFAR-10} & 0                    & -                      & -                     & -                     & 2.57                  \\
                          & 2                    & -                      & -                     & -                     & 12.57                 \\
                          & 4                    & -                      & 2.93                  & 9.32                  & 23.86                 \\ \hline
\end{tabular}
\end{sc}
\end{small}
\vspace{-0.15in}
\end{table}

\begin{table}[t]
\centering
\caption{Test accuracy improvement (\%) for \alg compared with FedAvg on MNIST ($q=0.8$, $s=4$). `-' means ``no statistical difference within $2\%$ error bar''.}
\label{table:SAFL_vs_FedAvg_mnist}
\begin{small}
\begin{sc}
\begin{tabular}{ccccc}
\hline
\multirow{2}{*}{Dataset Size} & \multicolumn{4}{c}{non-i.i.d. index ($p$)} \\ \cline{2-5} 
                              & 10      & 5      & 2          & 1          \\ \hline
50                            & -       & -      & 4.82       & 16.65      \\
100                           & -       & -      & 6.87       & 20.26      \\
500                           & -       & -      & 9.16       & 29.82      \\
1000                          & -       & -      & 10.69      & 31.07      \\ \hline
\end{tabular}
\end{sc}
\end{small}
\vspace{-0.15in}
\end{table}

\textbf{2) Improvement of the \alg Algorithm under Incomplete Client Participation:}
The improvements of our \alg can be observed in two aspects: improved test accuracy and faster convergence rate.

{\em I. Improved test accuracy.}
In Table~\ref{table:SAFL_vs_FedAvg} and ~\ref{table:SAFL_vs_FedAvg_mnist}, we show the test accuracy improvement of our \alg algorithm compared with that of FedAvg in standard FL.
In Table~\ref{table:SAFL_vs_FedAvg}, we can observe that even with a few server participation with a probability of $0.2$, there is a non-negligible improvement in test accuracy.
In Table~\ref{table:SAFL_vs_FedAvg_mnist},
the key observation is that, with a {\em small amount} of auxiliary data at the server, there is a significant increase of test accuracy for our \alg algorithm.
For example, with only $50$ data samples at the server ($0.1\%$ of the total training data), there is a $16.65\%$ test accuracy increase.
With $1000$ data samples, the improvement reaches $31.07\%$.
This verifies the effectiveness of our SA-FL framework and our \alg algorithm.
Another observation is that for nearly homogeneous case (e.g., from $p=10$ to $p=5$), there is no statistical difference with or without auxiliary data at the server (denoted by `-' in Table~\ref{table:SAFL_vs_FedAvg_mnist}.
This is consistent with the previous observations of negligible degradation in cases with homogeneous data across clients.

\begin{figure}[t]
    \centering    \label{fig:cifar10_SampleBias2_noniid1}\includegraphics[width=.4\textwidth]{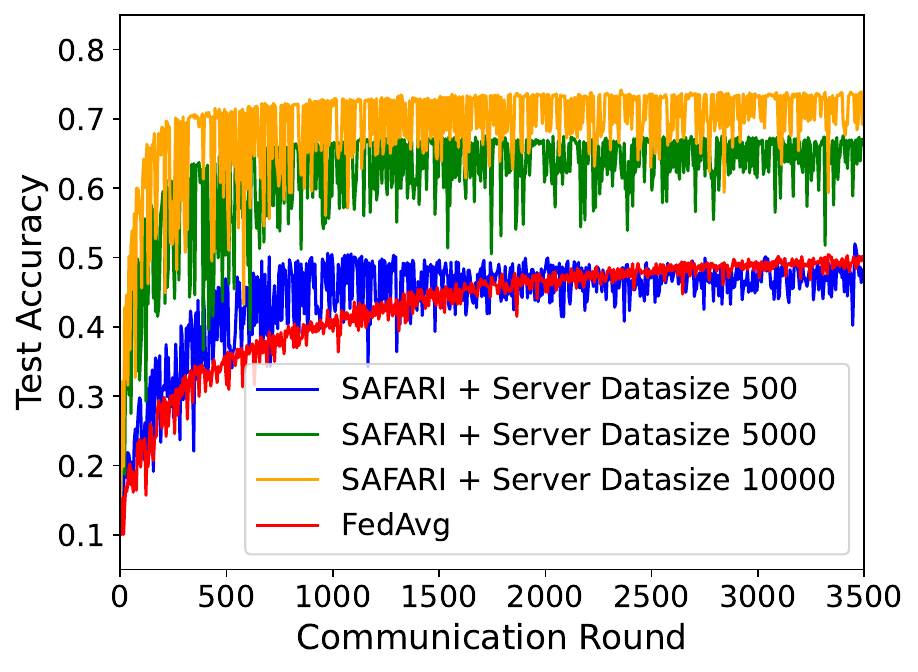}
    \vspace{-.1in}
    \caption{Comparison of test accuracy on CIFAR-10 ($s=4$, $p=1$, $q=0.4$).}
    \label{fig:cifar10_noniid1}
    \vspace{-.1in}
\end{figure}

{\em II. Faster convergence rate.}
In Fig.~\ref{fig:cifar10_noniid1}, we show the convergence processes of \alg on CIFAR-10 under incomplete client participation ($s=4$) with non-i.i.d. data ($p=1$).
We can see clearly that the convergence of \alg is accelerating and the test accuracy increases as more data are employed at the server.



\section{Conclusion} \label{sec: Conclusion}
In this paper, we rigorously investigated the server-assisted federated learning (SA-FL) framework (i.e., to deploy an auxiliary dataset at the server), which has been increasingly adopted in practice to mitigate the impacts of incomplete client participation in conventional FL.
To characterize the benefits of SA-FL, we first showed that conventional FL is {\em not} PAC-learnable under incomplete client participation by establishing a fundamental generalization error lower bound.
Then, we showed that SA-FL is able to revive the PAC-learnability of conventional FL under incomplete client participation.
Upon resolving the PAC-learnability challenge, we proposed a new \alg (\ul{s}erver-\ul{a}ssisted \ul{f}ederated \ul{a}ve\ul{r}ag\ul{i}ng) algorithm that enjoys convergence guarantee and the same level of communication efficiency as that of conventional FL.
Extensive numerical results also validated our theoretical findings.

\section*{Acknowledgements}
This work is supported in part by AI Seed Funding and the GWBC Award at
RIT, as well as NSF grants CAREER CNS-2110259, CNS-2112471, and IIS-2324052. 
The authors also thank Mr. Zhe Li for his assistance with the part of the experiments.

\section*{Impact Statement}

This paper delves into the realm of federated learning, a cutting-edge paradigm in machine learning that leverages decentralized training across multiple devices while preserving data privacy. The primary objective of our research is to contribute to the advancement of federated learning methodologies, addressing challenges and exploring opportunities for enhancing model performance in a privacy-preserving manner. 
As with any technological innovation, our work on federated learning has many potential societal consequences, none which we feel must be specifically highlighted here.


\bibliography{BIB/FederatedLearning, BIB/StatisticalLearning, BIB/Teams, BIB/Experiment}
\bibliographystyle{icml2024}

\newpage
\appendix
\onecolumn

\allowdisplaybreaks

\newpage
\section{Proofs} \label{sec: Proof}

\FLLB*
\begin{proof}
    Denote $S$ the dataset with size $Mn$ i.i.d. sampled from distribution $P$, $\mc{F}(\cdot)$ the sampling process of FL system, and $\bar{S} = \mc{F}(S)$ the training dataset selected by FL system with size $mn$.
    Consider a distribution $P$ with support on only two points $\{ x_1, x_2 \}$ such that $\mb{P}_{P} (x_1) = 1 - 4 \epsilon \ \text{and} \ \mb{P}_{P} (x_2) = 4 \epsilon$ with $\epsilon = \frac{1-\omega }{8}$.

    First we show that the rare points $x_2$ appears at most $(1 - \omega) Mn$ times with constant probability.
    Let $\hat{s}$ be the number of $x_2$ points in $S$, then $\hat{s} \sim \mb{B}(Mn, \epsilon)$ is a binomial random variable.
    By the Chernoff bound, 
    $$\mb{P} [\hat{s} \geq (1 - \omega) Mn] = \mb{P} [\hat{s} \geq (1 + 1) 4 \epsilon Mn] \leq e^{- \frac{4 \epsilon Mn}{3}} = e^{- \frac{(1 - \omega) Mn}{6}} \leq e^{- \frac{1}{6}} \leq \frac{17}{20} .$$
    So $\mb{P} [\hat{s} < (1 - \omega) Mn] > \frac{3}{20}$.

    Next, we consider the following sampling process with dataset $S = \{ (x^{'}_1, f(x^{'}_1)), \ldots , (x^{'}_{M \times n}, f(x^{'}_{M \times n})) \}$: choosing as many data $(x^{'}_i, f(x^{'}_i)), i \in [mn]$ such that $x^{'}_i = x_1$ as possible to form the training set $\bar{S}$.
    Let $f_1, f_2 \in \mc{H}$ be two target functions whose existence is guaranteed by the non-trivial definition of $\mc{H}$ and $f_1(x_1) = f_2(x_1), f_1(x_2) = - f_2(x_2)$, and $\mc{S}$ be the set of all datasets in $(\mc{X}, \mc{Y})^{(M \times n)}$ such that $\hat{s} < (1 - \omega) MN$.

    Let $\mc{R}(h_s, f) = \mb{P}_{P} [\mc{L}(\mc{F}(S))(x) \neq f_1(x) \cap x \neq x_1]$, the following holds for these two target functions $f_1$ and $f_2$:
    \begin{align}
    \mc{R}(h_s, f_1) + \mc{R}(h_s, f_2) 
    &= \mb{P}_{P} [\mc{L}(\mc{F}(S))(x) \neq f_1(x) \cap x \neq x_1] + \mb{P}_{P} [\mc{L}(\mc{F}(S))(x) \neq f_2(x) \cap x \neq x_1] \\
    &= \1_{\mc{L}(\mc{F}(S))(x_1) \neq f_1(x_1)} \mb{P}(x_2) + \1_{\mc{L}(\mc{F}(S))(x_1) \neq f_2(x_2)} \mb{P}(x_1) \\
    &= 4 \epsilon.
    \end{align}
    The above result hold in expectation since it holds for any $S \in \mc{S}$.
    Hence, there exists a target function $f \in \mc{H}$ such that $\mb{E}_{S \in \mc{S}} \mc{R}(h_s, f) \geq 2 \epsilon.$
    Note $\mc{R}(h_s, f) \leq \mb{P}(x \neq x_1) = 4 \epsilon$, then by decomposing the expectation into two parts we obtain:
    \begin{align}
        2 \epsilon \leq \mb{E}_{S \in \mc{S}} \mc{R}(h_s, f) 
        &= \sum_{S: \mc{R}(h_s, f) \geq \epsilon} \mc{R}(h_s, f) \mb{P} [\mc{R}(h_s, f)] + \sum_{S: \mc{R}(h_s, f) < \epsilon} \mc{R}(\mc{R}(h_s, f) \mb{P} [\mc{R}(h_s, f)] \\
        &\leq 4 \epsilon \mb{P}_{S \in \mc{S}} [\mc{R}(h_s, f) \geq 4 \epsilon] + \epsilon (1 - \mb{P}_{S \in \mc{S}} [\mc{R}(h_s, f) \geq \epsilon]) \\
        &= \epsilon + 3 \epsilon \mb{P}_{S \in \mc{S}} [\mc{R}(h_s, f) \geq \epsilon].
    \end{align}
    That is,
    \begin{align}
        \mb{P}_{S \in \mc{S}} [\mc{R}(h_s, f) \geq \epsilon] \geq \frac{1}{3}.
    \end{align}
    Note $\mc{R}(h_s, f) = \mb{P}_{P} [\mc{L}(\mc{F}(S))(x) \neq f_1(x) \cap x \neq x_1] \leq \mc{R}(\mc{L}(\mc{F}(S))) = \mb{P}_{P} [\mc{L}(\mc{F}(S))(x) \neq f_1(x)]$, then we have the final results:
    \begin{align}
        \mb{P}_{S \sim P} [\mc{R}_P(\mc{L}(\mc{F}(S)), f) \geq \epsilon] 
        &\geq \mb{P}_{S \sim P} [\mc{R}(h_s, f) \geq \epsilon] \\
        &\geq \mb{P}_{S \in \mc{S}} [\mc{R}(h_s, f) \geq \epsilon] \mb{P} [S \in \mc{S}] \\
        &> \frac{1}{3} \frac{3}{20} = \frac{1}{20}.
    \end{align}
 \end{proof}

\FALbound*
\begin{proof}
    \begin{align}
        \varepsilon_{P}(\hat{h}_Q^{*}) 
        &= \mc{R}_{P}(\hat{h}_Q^{*}) - \mc{R}_{P}(h_{P}^{*}) \\
        &= [\mc{R}_{P}(\hat{h}_Q^{*}) - \mc{R}_{P}(h_{P}^{*}) - (\mc{R}_{Q}(\hat{h}_Q^{*}) - \mc{R}_{Q}(h_Q^{*}))] + \mc{R}_{Q}(\hat{h}_Q^{*}) - \mc{R}_{Q}(h_Q^{*}) \\
        &\leq | \varepsilon_{P}(\hat{h}_Q^{*}) - \varepsilon_{Q}(\hat{h}_Q^{*}) | + \varepsilon_{Q}(\hat{h}_Q^{*}) \\
        &\leq \alpha \varepsilon_{Q}(\hat{h}_Q^{*})^\beta + \varepsilon_{Q}(\hat{h}_Q^{*}).
    \end{align}
    
    Combining with Lemma~\ref{lm:AuxiliaryLemma}, the proof is complete.
    
    \begin{restatable} [Auxiliary Lemma~\citep{massart2006risk,koltchinskii2006local,hanneke2019value,hanneke2020no}] {lemma} {AuxiliaryLemma}
        \label{lm:AuxiliaryLemma}
            For any $m \in \mb{N}$ and $\delta \in (0, 1)$, define $A(m, \delta)=\frac{d_{\mc{H}}}{m} \log(\frac{m}{d_{\mc{H}}} + \frac{1}{m} \log(\frac{1}{\delta}))$ 
            With probability at least $1-\delta$, $\forall h, \hat{h} \in \mc{H}$,
            \begin{align}
                &\mc{R}(h) - \mc{R}(\hat{h}) \leq \mc{\hat{R}}(h) - \mc{\hat{R}}(\hat{h}) + c \sqrt{ \min{ \{ \mb{P}_S(h \neq \hat{h}), \hat{\mb{P}}_S(h \neq \hat{h}) \} } A(m, \delta)} + c A(m, \delta), \\
                &\frac{1}{2} \mb{P}_S(h \neq \hat{h}) - c A(m, \delta) \leq \hat{\mb{P}}_S(h \neq \hat{h}) \leq 2 \mb{P}_S(h \neq \hat{h}) + c A(m, \delta), \\
                &\varepsilon_{Q}(\hat{h}_Q^{*}) = \left[ A(m, \delta) \right] ^{\frac{1}{2-\beta_Q}},
            \end{align}
            where $\mb{P}_S(\cdot) = \mb{E} [\hat{\mb{P}}_S(\cdot)]$, $S$ is the i.i.d. dataset with size $m$ drawn form distribution $Q$, $c \in (0, \infty)$ is a constant.
    \end{restatable}
    
\end{proof}

\FALCL*
\begin{proof}
Without loss of generality, we use $c$ serve as a generic constant since we focus on the order in terms of the sample number and thus omit the constant factor.

    \begin{align}
        \varepsilon_{P}(\hat{h}_Q^*) &= \mc{R}_P(\hat{h}_Q^*) - \mc{R}_P(h_P^*) \\
        &\leq \mc{\hat{R}}_P(\hat{h}_Q^*) - \mc{\hat{R}}_P(h_P^*) + c \sqrt{ \min{ \{ P(\hat{h}_Q^* \neq h_P^*), \hat{P}(\hat{h}_Q^* \neq h_P^*) \} } A(n_T, \delta)} + c A(n_T, \delta) \\
        &\leq c \sqrt{ \varepsilon_{P}^{\beta_P} (\hat{h}_Q^*) A(n_T, \delta)} + c A(n_T, \delta).
    \end{align}

    The first inequality is due to Lemma~\ref{lm:AuxiliaryLemma} and second inequality follows from Lemma~\ref{lm:Lemma_appx} and Noise assumption~\ref{assum: NC}.
    Then we have the following result, which completes the proof:

    \begin{eqnarray*}
        \varepsilon_{P}(\hat{h}_Q^*) \leq c A(n_T, \delta)^{\frac{1}{2-\beta_P}}.
    \end{eqnarray*}
\end{proof}

\begin{restatable} [] {lemma} {Lemma_appx}
    \label{lm:Lemma_appx}
    If $\mc{\hat{R}}_P(\hat{h}_Q^*) \leq \mc{\hat{R}}_P(h_Q^*)$, 
        with probability at least $1-\delta$,
        \begin{eqnarray*}
            \mc{\hat{R}}_P(\hat{h}_Q^*) - \mc{\hat{R}}_P(h_P^*) = \varepsilon_{P}(h_Q^*) + \mc{O}(A(n_T, \delta)).
        \end{eqnarray*}
\end{restatable}

\begin{proof}
\begin{align}
    &\mc{\hat{R}}_P(\hat{h}_Q^*) - \mc{\hat{R}}_P(h_P^*) \leq \mc{\hat{R}}_P(h_Q^*) - \mc{\hat{R}}_P(h_P^*) \\
    &\leq \mc{R}_P(h_Q^*) - \mc{R}_P(h_P^*) + c \sqrt{ \min{ \{ P(h_Q^* \neq h_P^*), \hat{P}(h_Q^* \neq h_P^*) \} } A(n_T, \delta)} + c A(n_T, \delta) \\
    &= \varepsilon_{P}(h_Q^*) + \mc{O}(A(n_T, \delta)).
\end{align}
\end{proof}

\FAA*

\begin{proof}
In expectation, we define that there are totally $R_s = | \mathcal{T}_s | = (1-p) R$ rounds for server update, $R_c = | \mathcal{T}_c | = p R$ rounds for client update, and $R = R_s + R_c$, 

Taking expectation on the random data samples conditioned on $\x_r$, we can have the following one-step descent when server updates:
\begin{align}
    &\mathbb{E}_r [F(\x_{r+1})] \leq F(\x_r) + \big< \nabla F(\x_r), \mathbb{E}_r [\x_{r+1} - \x_r] \big> + \frac{L}{2} \mathbb{E}_r [\| \x_{r+1} - \x_r \|^2] \\
    &= F(\x_r) + \big< \nabla F(\x_r), \eta_s \mathbb{E}_r [\nabla F(\x_r, \xi_r)] \big> + \frac{L}{2} \eta_s^2 \mathbb{E}_r [ \| \nabla F(\x_r, \xi_r) \|^2 ]\\
    &= F(\x_r) - \eta_s \| \nabla F(\x_r) \|^2 + \frac{L \eta_s^2}{2} \| \nabla F(\x_r) \|^2 + \frac{L \eta_s^2}{2} \sigma_s^2.
\end{align}
That is,
\begin{align}
    &\| \nabla F(\x_r) \|^2 \leq \frac{2}{\eta_s} (F(\x_r) - \mathbb{E}_r[F(\x_{r+1})]) + (L \eta_s - 1) \| \nabla F(\x_r) \|^2 + L \eta_s \sigma_s^2.
\end{align}

Similarly, when clients update, we assume there are totally $m$ clients participating in one round, denoted as $[m]$. Then we have:
\begin{align}
    \mathbb{E}_r [F(\x_{r+1})] &\leq F(\x_r) + \big< \nabla F(\x_r), \mathbb{E}_r [\x_{r+1} - \x_r] \big> + \frac{L}{2} \mathbb{E}_r [\| \x_{r+1} - \x_r \|^2] \\
    &= F(\x_r) + \underbrace{\big< \nabla F(\x_r), - \eta_c \mathbb{E}_r [\Delta_r] \big>}_{A_1} + \underbrace{ \frac{L}{2} \eta_c^2 \mathbb{E}_r [ \| \Delta_r  \|^2 ]}_{A_2}.
\end{align}

\begin{align}
    A_1 &= \big< \nabla F(\x_r), - \eta_c \mathbb{E}_r [\Delta_r] \big> \\
    &= \frac{1}{2K} \eta_c \left[ - K^2 \| \nabla F(\x_r) \|^2 - \| \mathbb{E}_r [\Delta_r] \|^2 + \| K \nabla F(\x_r) - \mathbb{E}_r [\Delta_r] \|^2 \right] \\
    &= - \frac{K \eta_c}{2} \| \nabla F(\x_r) \|^2 - \frac{\eta_c}{2K} \left\| \frac{1}{m} \sum_{i \in S_r} \sum_{k \in [K]} \nabla F_i(\x_{r, k}^i) \right\|^2 + \frac{\eta_c}{2K} \left\| \frac{1}{m} \sum_{i \in S_r} \sum_{k \in [K]} \left[ \nabla F(\x_r) - \nabla F_i(\x_{r, k}^i) \right] \right\|^2 \\
    &\leq - \frac{K \eta_c}{2} \| \nabla F(\x_r) \|^2 - \frac{\eta_c}{2K} \left\| \frac{1}{m} \sum_{i \in S_r} \sum_{k \in [K]} \nabla F_i(\x_{r, k}^i) \right\|^2 + \frac{\eta_c}{2m} \sum_{i \in S_r} \sum_{k \in [K]} \underbrace{\left\| \nabla F(\x_r) - \nabla F_i(\x_{r, k}^i) \right\|^2}_{A_3} \\
    &\leq - \frac{K \eta_c}{2} \| \nabla F(\x_r) \|^2 - \frac{\eta_c}{2K} \left\| \frac{1}{m} \sum_{i \in S_r} \sum_{k \in [K]} \nabla F_i(\x_{r, k}^i) \right\|^2 \\
    & \quad + \eta_c K \sigma_G^2 + \eta_c K L^2 \left[5K \eta_c^2 (\sigma^2 + 6K \sigma_G^2) + 30K^2 \eta_c^2 \| \nabla F(\x_r) \|^2 \right],
\end{align}
where $A_3$ could be bounded as follows:
\begin{align}
    A_3 &= \left\| \nabla F(\x_r) - \nabla F_i(\x_{r, k}^i) \right\|^2 \\
    &= \left\| \nabla F(\x_r) - \nabla F_i(\x_r) + \nabla F_i(\x_r) - \nabla F_i(\x_{r, k}^i) \right\|^2 \\
    &\leq 2 \left\| \nabla F(\x_r) - \nabla F_i(\x_r) \right\|^2 + 2 \left\| \nabla F_i(\x_r) - \nabla F_i(\x_{r, k}^i) \right\|^2 \\
    &\leq 2 \sigma_G^2 + 2 L^2 \left\| \x_r - \x_{r, k}^i \right\|^2 \\
    &\leq 2 \sigma_G^2 + 2 L^2 \left[5K \eta_c^2 (\sigma^2 + 6K \sigma_G^2) + 30K^2 \eta_c^2 \| \nabla F(\x_r) \|^2 \right],
\end{align}
where the last inequality follows from the bounded local update step with $\eta_c \leq \frac{1}{8LK}$ (see Lemma 2 in \citep{yang2021linearspeedup} and Lemma 3 in \citep{reddi2021adaptive}).

    \begin{align}
        A_2 &= \frac{L}{2} \eta_c^2 \mathbb{E}_r [ \| \Delta_r  \|^2 ] \\
        &\leq L \eta_c^2 \left\| \frac{1}{m} \sum_{i \in S_r} \sum_{k \in [K]} \nabla F_i(\x_{r, k}^i) \right\|^2 + L \eta_c^2 \left\| \frac{1}{m} \sum_{i \in S_r} \sum_{k \in [K]} \left[ \nabla F_i(\x_{r, k}^i) - \nabla F_i(\x_{r, k}^i, \xi_{r, k}^i) \right] \right\|^2 \\
        &\leq L \eta_c^2 \left\| \frac{1}{m} \sum_{i \in S_r} \sum_{k \in [K]} \nabla F_i(\x_{r, k}^i) \right\|^2 + \frac{L \eta_c^2 K}{m} \sigma^2,
    \end{align}
    where the last inequality is due to the martingale difference sequence $\{ \nabla F_i(\x_{r, k}^i) - \nabla F_i(\x_{r, k}^i, \xi_{r, k}^i) \}$ (see Lemma 4 in~\citep{Karimireddy2020SCAFFOLD}).

    Putting pieces together, we have
    \begin{align}
        & K \eta_c (\frac{1}{2} - 30 L^2 K^2 \eta_c^2) \| \nabla F(\x_r) \|^2 \leq F(\x_r) - \mathbb{E}_r[F(\x_{r+1}) - \frac{\eta_c}{2K} \left\| \frac{1}{m} \sum_{i \in S_r} \sum_{k \in [K]} \nabla F_i(\x_{r, k}^i) \right\|^2 + \eta_c K \sigma_G^2 \\
        &\quad + \eta_c K L^2 \left[5K \eta_c^2 (\sigma^2 + 6K \sigma_G^2) \right] + L \eta_c^2 \left\| \frac{1}{m} \sum_{i \in S_r} \sum_{k \in [K]} \nabla F_i(\x_{r, k}^i) \right\|^2 + \frac{L \eta_c^2 K}{m} \sigma^2
    \end{align}

    If $(\frac{1}{2} - 30 L^2 K^2 \eta_c^2) \geq \frac{1}{4}$ (i.e., $\eta_c \leq \frac{1}{4\sqrt{30} LK}$) and $\eta_c = \frac{2 \eta_s}{K}$, we have
    \begin{align}
        &\| \nabla F(\x_r) \|^2 \leq \frac{4}{K \eta_c} (F(\x_r) - \mathbb{E}_r[F(\x_{r+1})]) + 4 \sigma_G^2 \\
        &\quad + (\frac{4L\eta_c}{K} - \frac{2}{K^2}) \left\| \frac{1}{m} \sum_{i \in S_r} \sum_{k \in [K]} \nabla F_i(\x_{r, k}^i) \right\|^2 + 20 K L^2 \eta_c^2(\sigma^2 + 6K \sigma_G^2) + \frac{4 L \eta_c}{m} \sigma^2 \\
        &= \frac{2}{\eta_s} (F(\x_r) - \mathbb{E}_r[F(\x_{r+1})]) + 4 \sigma_G^2 \\
        &\quad + (\frac{8L\eta_s}{K^2} - \frac{2}{K^2}) \left\| \frac{1}{m} \sum_{i \in S_r} \sum_{k \in [K]} \nabla F_i(\x_{r, k}^i) \right\|^2 + \frac{80 L^2 \eta_s^2}{K}(\sigma^2 + 6K \sigma_G^2) + \frac{8L \eta_s}{mK} \sigma^2
    \end{align}

    Note there are totally $R_s$ rounds ($T_s$ as the round indices) for server update and $R_c$ rounds ($T_c$ as the round indices) for client update. Let $R = R_s + R_c$, we have 
    \begin{align}
        & \frac{1}{R} \sum_{r = 1}^R \mathbb{E} \| \nabla F(\x_r) \|^2 \leq \frac{2}{\eta_s} \frac{1}{R} \sum_{r = 1}^R (F(\x_r) - F(\x_{r+1})) + \frac{1}{R} \sum_{r \in T_s} (L \eta_s - 1) \| \nabla F(\x_r) \|^2 + \frac{L \eta_s R_s}{R} \sigma_s^2 \\
        &\quad + \frac{4 R_c}{R} \sigma_G^2 + \frac{1}{R} \sum_{r \in T_c} (\frac{8L\eta_s}{K^2} - \frac{2}{K^2}) \left\| \frac{1}{m} \sum_{i \in S_r} \sum_{k \in [K]} \nabla F_i(\x_{r, k}^i) \right\|^2 + \frac{80 R_c L^2 \eta_s^2}{K R}(\sigma^2 + 6K \sigma_G^2) + \frac{8L R_c \eta_s}{mKR} \sigma^2 \\
        &\leq \frac{2 (F(\x_0) - F(\x^*))}{R \eta_s} + \frac{L \eta_s R_s}{R} \sigma_s^2 + \frac{80 R_c L^2 \eta_s^2}{K R}(\sigma^2 + 6K \sigma_G^2) + \frac{8L R_c \eta_s}{mKR} \sigma^2, 
    \end{align}
    where the last inequality follows from 
    \begin{align}
        4 R_c \sigma_G^2 &\leq \sum_{r \in T_s} (1- L \eta_s) \| \nabla F(\x_r) \|^2 + \sum_{r \in T_c} (\frac{2}{K^2} - \frac{8L\eta_s}{K^2}) \left\| \frac{1}{m} \sum_{i \in S_r} \sum_{k \in [K]} \nabla F_i(\x_{r, k}^i) \right\|^2 \\
        &\leq R_s (1- L \eta_s) G_1 + R_c (\frac{2}{K^2} - \frac{8L\eta_s}{K^2}) G_2,
    \end{align}
    where $G_1 = \max_{r \in T_s} \| \nabla F(\x_r) \|^2$ and $G_2 = \max_{r \in T_c} \left\| \frac{1}{m} \sum_{i \in S_r} \sum_{k \in [K]} \nabla F_i(\x_{r, k}^i) \right\|^2$.
    That is the requirement on $q$ such that $q \leq 1 / \left( \frac{4 \sigma_G^2 - 4G_2 (\frac{1}{2K^2} - \frac{2L \eta_s^2}{K^2})}{(1 - L \eta_s) G_1} + 1 \right)$.

\end{proof}

\subsection{Discussions} \label{sebsec:discussions}
We want to cast caveats on Corollary~\ref{cor:convergence_FAA}. The results in Corollary~\ref{cor:convergence_FAA} does not hold in {\em arbitrary} cases. Specifically, Corollary~\ref{cor:convergence_FAA} requires both $R_s \geq \frac{4 \sigma_G^2 - 4G_2 (\frac{1}{2K^2} - \frac{2L \eta_s^2}{K^2})}{(1 - L \eta_s) G_1} R_c$ and $R_s = \mathcal{O}(\frac{R}{mK})$.
In other words, we need $\frac{4 \sigma_G^2 - 4G_2 (\frac{1}{2K^2} - \frac{2L \eta_s^2}{K^2})}{(1 - L \eta_s) G_1} \leq \frac{R_s}{R_c} \leq \frac{c}{mK(1 - \frac{c}{mK})}$, where $c$ is a constant.
Approximately, $\frac{R_s}{R_c} = constant$. Due to $R_s + R_c = R$, we can see that $R_c = \Omega(R)$. So the convergence rate is $\mathcal{O} ( \frac{1}{\sqrt{mKR}} )$, which is the same order of $\mathcal{O} ( \frac{1}{\sqrt{mKR_c}} )$. 

\FAS*

\begin{proof}
Taking expectation on the random data samples conditioned on $\x_r$, we can have the following one-step descent as classic stochastic gradient descent method when server updates:
\begin{align}
    \mathbb{E}_r \| \x_{r+1} - \x_* \|^2 &= \mathbb{E}_r \| \x_r - \eta_s \nabla F(\x_r, \xi_r) - \x_* \|^2 \\
    &\leq \| \x_r - \x_* \|^2 + \eta_s^2 \| \nabla F(\x_r) \|^2 + \eta_s^2 \sigma_s^2 - 2 \eta_s \big< \x_r - \x_* , \nabla F(\x_r) \big>\\
    &\leq \left[ 1 - \frac{2 \eta_s L\mu}{L + \mu} \right] \| \x_r - \x_* \|^2 + \eta_s (\eta_s - \frac{2}{L + \mu} ) \| \nabla F(\x_r) \|^2 + \eta_s^2 \sigma_s^2 
\end{align}
where the second last inequality is due to the $\mu-$strongly convex property (see Lemma 3.11 in \citep{bubeck2015convex}) and the last inequality follows from the fact $\eta_s \leq \frac{2}{L + \mu}$ and $\mu-$strongly convex property $\| \nabla F(\x) \|^2 \geq 2 \mu (F(\x) - F(\x_*))$.

When arbitrary clients participate, we have
\begin{align}
    \mathbb{E}_r \| \x_{r+1} - \x_* \|^2 &= \mathbb{E}_r \| \x_r - \eta_c \Delta_r - \x_* \|^2 \\
    &\leq \| \x_r - \x_* \|^2 + \eta_c^2 \mathbb{E}_r \left\| \frac{1}{m} \sum_{i \in S_r} \sum_{k \in [K]} \nabla F_i(\x_{r, k}^i, \xi_{r, k}^i) \right\|^2 - 2 \eta_c \big< \x_r - \x_* , \frac{1}{m} \sum_{i \in S_r} \sum_{k \in [K]} \nabla F_i(\x_{r, k}^i) \big> \\
    &\leq \| \x_r - \x_* \|^2 + \eta_c^2 \mathbb{E}_r \left\| \frac{1}{m} \sum_{i \in S_r} \sum_{k \in [K]} \nabla F_i(\x_{r, k}^i) \right\|^2 + \frac{K \eta_c^2}{m} \sigma^2 - 2 \eta_c \big< \x_r - \x_* , \frac{1}{m} \sum_{i \in S_r} \sum_{k \in [K]} \nabla F_i(\x_{r, k}^i) \big> \\
    &\leq \| \x_r - \x_* \|^2 + \eta_c^2 \mathbb{E}_r \left\| \frac{1}{m} \sum_{i \in S_r} \sum_{k \in [K]} \nabla F_i(\x_{r, k}^i) \right\|^2 + \frac{K \eta_c^2}{m} \sigma^2 \\
    & - 2 \eta_c \frac{1}{m} \sum_{i \in S_r} \sum_{k \in [K]} \left[ F_i(\x_r) - F_i(\x_*) + \frac{\mu}{4} \| \x_r - \x_* \|^2 - L \| \x_r - \x_{r, k}^i \|^2 \right] \\
    &\leq (1 - \frac{\eta_c K \mu}{2}) \| \x_r - \x_* \|^2 + \eta_c^2 \mathbb{E}_r \left\| \frac{1}{m} \sum_{i \in S_r} \sum_{k \in [K]} \nabla F_i(\x_{r, k}^i) \right\|^2 + \frac{K \eta_c^2}{m} \sigma^2 \\
    & - \frac{2 \eta_c K}{m} \sum_{i \in S_r} \left[ F_i(\x_r) - F_i(\x_*) \right] + \frac{2 \eta_c L}{m} \sum_{i \in S_r} \sum_{k \in [K]} \| \x_r - \x_{r, k}^i \|^2  \\
\end{align}
where the second last inequality is due to $\big< \nabla f(\x), \z - \y \big> \geq f(\z) - f(\y) + \frac{\mu}{4} \| \z - \y \|^2 - L \| \z -\x \|^2$ for any $\mu-$strongly convex and L-smooth function $f$ (see Lemma 5 in \citep{Karimireddy2020SCAFFOLD}).

\begin{align}
    & \mathbb{E}_r \left\| \frac{1}{m} \sum_{i \in S_r} \sum_{k \in [K]} \nabla F_i(\x_{r, k}^i) \right\|^2 = \mathbb{E}_r \left\| \frac{1}{m} \sum_{i \in S_r} \sum_{k \in [K]} \nabla F_i(\x_{r, k}^i) - \nabla F_i(\x_r) + \nabla F_i(\x_r) \right\|^2\\
    &\leq 2 \mathbb{E}_r \left\| \frac{1}{m} \sum_{i \in S_r} \sum_{k \in [K]} \nabla F_i(\x_{r, k}^i) - \nabla F_i(\x_r) \right\|^2 + 2 \mathbb{E}_r \left\| \frac{1}{m} \sum_{i \in S_r} \sum_{k \in [K]} \nabla F_i(\x_r) \right\|^2 \\
    &\leq \frac{2K}{m} \sum_{i \in S_r} \sum_{k \in [K]} \mathbb{E}_r \left\| \nabla F_i(\x_{r, k}^i) - \nabla F_i(\x_r) \right\|^2 + 4 \mathbb{E}_r \left\| \frac{1}{m} \sum_{i \in S_r} \sum_{k \in [K]} \nabla F_i(\x_r) - \nabla F(\x_r) \right\|^2 + 4 K^2 \mathbb{E}_r \left\| \nabla F(\x_r) \right\|^2 \\
    &\leq \frac{2KL^2}{m} \sum_{i \in S_r} \sum_{k \in [K]} \mathbb{E}_r \left\| \x_{r, k}^i - \x_r \right\|^2 + 4 K^2 \sigma_G^2 + 4 K^2 \mathbb{E}_r \left\| \nabla F(\x_r) \right\|^2 \\
\end{align}

Then we have

\begin{align}
    \mathbb{E}_r \| \x_{r+1} - \x_* \|^2 &\leq (1 - \frac{\eta_c K \mu}{2}) \| \x_r - \x_* \|^2 + 4 \eta_c^2 K^2 \sigma_G^2 + 4 \eta_c^2 K^2 \left\| \nabla F(\x_r) \right\|^2 + \frac{K \eta_c^2}{m} \sigma^2 \\
    & - \frac{2 \eta_c K}{m} \sum_{i \in S_r} \left[ F_i(\x_r) - F_i(\x_*) \right] + \frac{2 \eta_c L(1 + K \eta_c L)}{m} \sum_{i \in S_r} \sum_{k \in [K]} \mathbb{E}_r \| \x_r - \x_{r, k}^i \|^2  \\
    &\leq (1 - \frac{\eta_c K \mu}{2}) \| \x_r - \x_* \|^2 + 4 \eta_c^2 K^2 \sigma_G^2 + 4 \eta_c^2 K^2 \left\| \nabla F(\x_r) \right\|^2 + \frac{K \eta_c^2}{m} \sigma^2 \\
    & - \frac{2 \eta_c K}{m} \sum_{i \in S_r} \left[ F_i(\x_r) - F_i(\x_*) \right] + 2 \eta_c K L(1 + K \eta_c L) \left[5K \eta_c^2 (\sigma^2 + 6K \sigma_G^2) + 30K^2 \eta_c^2 \| \nabla F(\x_r) \|^2 \right] \\
    &= (1 - \frac{\eta_c K \mu}{2}) \| \x_r - \x_* \|^2 + 4 \eta_c^2 K^2 \sigma_G^2 + 4 \eta_c^2 K^2 [1 + 15 \eta_c K L (1+ \eta_c K L)] \left\| \nabla F(\x_r) \right\|^2 + \frac{K \eta_c^2}{m} \sigma^2 \\
    & - \frac{2 \eta_c K}{m} \sum_{i \in S_r} \left[ F_i(\x_r) - F_i(\x_*) \right] + 2 \eta_c K L(1 + K \eta_c L) \left[5K \eta_c^2 (\sigma^2 + 6K \sigma_G^2) \right].
\end{align}

The last inequality is due to the upper bound of $\mathbb{E}_r \| \x_r - \x_{r, k}^i \|^2$. That is, for each client $i \in [M]$, we have $\mathbb{E}_r \| \x_r - \x_{r, k}^i \|^2 \leq 5K \eta_c^2 (\sigma^2 + 6K \sigma_G^2) + 30K^2 \eta_c^2 \| \nabla F(\x_r) \|^2.$

For each step, we choose to use clients' update with probability $q$ and server's update with probability $1-q$. 
Taking expectation and letting $\bar{\eta} = \frac{\eta_c K \mu}{2} = \frac{2 \eta_s L\mu}{L + \mu},$ we have the following:
\begin{align}
    &\mathbb{E} \| \x_{r+1} - \x_* \|^2 
    \leq (1 - \bar{\eta}) \| \x_r - \x_* \|^2 + 4 q \eta_c^2 K^2 \sigma_G^2 + 4 q \eta_c^2 K^2 [1 + 15 \eta_c K L (1+ \eta_c K L)] \left\| \nabla F(\x_r) \right\|^2 + \frac{q K \eta_c^2}{m} \sigma^2 \\
    & - \frac{2 q \eta_c K}{m} \sum_{i \in S_r} \left[ F_i(\x_r) - F_i(\x_*) \right] + 2 q  \eta_c K L(1 + K \eta_c L) \left[5K \eta_c^2 (\sigma^2 + 6K \sigma_G^2) \right] \\
    & + (1 - q) \eta_s \left(\eta_s - \frac{2}{L + \mu} \right) \| \nabla F(\x_r) \|^2 + (1 - q) \eta_s^2 \sigma_s^2 \\
    &= (1 - \bar{\eta}) \| \x_r - \x_* \|^2 + \bar{\eta} \left[ \frac{4q \bar{\eta}}{\mu^2} \left(1 + \frac{30 L \bar{\eta}}{\mu}(1 + \frac{2L \bar{\eta}}{\mu}) \right) + (1-q) \frac{L + \mu}{2L \mu} \left( \frac{(L + \mu) \bar{\eta}}{2L \mu} - \frac{2}{L + \mu} \right) \right] \left\| \nabla F(\x_r) \right\|^2 
    \\
    & - \frac{4q \bar{\eta}}{ \mu m} \sum_{i \in S_r} \left[ F_i(\x_r) - F_i(\x_*) \right] + 4 q \eta_c^2 K^2 \sigma_G^2  + \frac{q K \eta_c^2}{m} \sigma^2 + 2 q  \eta_c K L(1 + K \eta_c L) \left[5K \eta_c^2 (\sigma^2 + 6K \sigma_G^2) \right] + (1 - q) \eta_s^2 \sigma_s^2 \\
    &\leq (1 - \bar{\eta}) \| \x_r - \x_* \|^2 + 4 q \eta_c^2 K^2 \sigma_G^2  + \frac{q K \eta_c^2}{m} \sigma^2 + 2 q  \eta_c K L(1 + K \eta_c L) \left[5K \eta_c^2 (\sigma^2 + 6K \sigma_G^2) \right] + (1 - q) \eta_s^2 \sigma_s^2,
\end{align}
where $\left[ \frac{4q \bar{\eta}}{\mu^2} \left(1 + \frac{30 L \bar{\eta}}{\mu}(1 + \frac{2L \bar{\eta}}{\mu}) \right) + (1-q) \frac{L + \mu}{2L \mu} \left( \frac{(L + \mu) \bar{\eta}}{2L \mu} - \frac{2}{L + \mu} \right) \right] \left\| \nabla F(\x_r) \right\|^2 - \frac{4q}{ \mu m} \sum_{i \in S_r} \left[ F_i(\x_r) - F_i(\x_*) \right] \leq 0$. 
That is, $q \leq \frac{1}{1 + \frac{\frac{4\bar{\eta}}{\mu^2} \left(1 + \frac{30 L \bar{\eta}}{\mu}(1 + \frac{2L \bar{\eta}}{\mu}) \right) G_3 - \frac{4}{ \mu} G_4}{\left(\frac{1}{L + \mu} - \frac{(L+\mu)^2 \bar{\eta}}{4L^2 \mu^2}\right) G_3 }}$, where $G_3 = \left\| \nabla F(\x_r) \right\|^2 $ and $G_4 = \frac{1}{m} \sum_{i \in S_r} \left[ F_i(\x_r) - F_i(\x_*) \right]$.

Recursively applying the above and summing up the geometric series gives:
\begin{align}
    \mathbb{E} \| \x_{R} - \x_* \|^2 
    &\leq (1 - \bar{\eta})^R \| \x_0 - \x_* \|^2 + \sum_{r=0}^{R-1} (1 - \bar{\eta})^j \delta \\
    &\leq (1 - \bar{\eta})^R \| \x_0 - \x_* \|^2 +  \bar{\delta},
\end{align}
where $\bar{\delta} = \frac{8 q \eta_c K}{\mu} \sigma_G^2  + \frac{2 q \eta_c}{\mu m} \sigma^2 + \frac{4 q  L(1 + K \eta_c L)}{\mu} \left[5K \eta_c^2 (\sigma^2 + 6K \sigma_G^2) \right] + (1 - q) \frac{L + \mu}{2 L \mu} \eta_s \sigma_s^2$.

Choosing $\bar{\eta} = \Omega(\frac{log(R)}{R})$, that is, $\eta_c = \Omega(\frac{log(R)}{R})$ and $\eta_s = \Omega(\frac{log(R)}{R})$, we have 
$$
\mathbb{E} \| \x_{R} - \x_* \|^2 \leq \tilde{\mathcal{O}}(\frac{1}{R}).
$$

\end{proof}

\section{Experiments} \label{sec: Experiments}

In this section, we provide the details of the numerical experiments and some additional experimental results.

\subsection{Models and Datasets}
We test the \alg algorithm by running two models on two different types of datasets, including 1) multinomial logistic regression (LR) on MNIST, and 2) convolutional neural network (CNN) on CIFAR-10. Both datasets are chose from a previous FL paper~\citep{mcmahan2017communication}, and they are now widely used as benchmarks for FL research~\citep{yang2021linearspeedup, Li2020convergence}.

MNIST and CIFAR-10 have ten classes of images separately. In order to impose the heterogeneity of the data, we partition the dataset according to the number of classes ($p$) that each client contains. We distribute these data to $M=10$ clients, and each client only has a certain number of classes. Specifically, each client randomly selects $p$ classes of images and then evenly samples training and test data-points within these $p$ classes of images without replacement. For example, if $p=2$, each client only samples training and test data-points within two classes of images, which causes the heterogeneity among different clients. If $p=10$, each client contains training and test samples that selects from ten classes. This situation is almost the same as i.i.d. case. Hence, the number of classes ($p$) in each client's local dataset can be used to represent the level of non-i.i.d. qualitatively. In addition, to mimic incomplete client participation, we enforce $s$ clients to be exempt from participation, where the index $s$ can be used to represent the degree of incomplete client participation. Specifically, we assume there are $M=10$ clients in total, and $m=5$ clients participate in each communication round. These clients are uniformly sampled from $M-s$ clients. Larger incomplete client participation index $s$ means less clients participate in the training.

For both MNIST and CIFAR-10, the global learning learning rate 1.0, the local learning rate is 0.1, and the server learning rate for \alg is 0.1. The local epoch is 1. For MNIST, the batch size is 64, and the total communication round is 150. For CIFAR-10, the batch size is 500, and the total communication round is 5000. To simulate the data heterogeneity, we use $p=[10, 5, 2, 1]$ as a proxy to represent the degree of non-i.i.d. on MNIST and CIFAR-10 datasets. To emulate the effect of incomplete client participation, we set $s=[0, 2, 4]$ to represent the degree of incomplete client participation for the \alg algorithm and the FedAvg algorithm. FedAvg is employed as the baselines to compare with our algorithm. To compare the effect of the collaboration from server, we add $[50, 100, 500, 1000]$ data to the server's side for MNIST and $[500, 5000, 10000]$ for CIFAR-10 and choose the client update probability $q=[0.4, 0.6, 0.8, 1.0]$.
In the case of $q=1.0$, our proposed algorithm \alg is equivalent to FedAvg.

\subsection{Additional Experimental Results}

\begin{table}[t!]
\caption{Test accuracy improvement (\%) for \alg compared with FedAvg on CIFAR-10 ($s=4$, $q=0.8$). `-' means no statistical difference within $2\%$ error bar.}
\label{table:SAFL_vs_FedAvg_cifar10}
\vskip 0.15in
\begin{center}
\begin{small}
\begin{sc}
\begin{tabular}{ccccc}
\hline
\multirow{2}{*}{Dataset Size} & \multicolumn{4}{c}{non-i.i.d. index ($p$)} \\ \cline{2-5} 
                              & 10      & 5         & 2        & 1         \\ \hline
500                           & -       & -         & 3.45     & 7.14      \\
5000                          & -       & -         & 8.15     & 20.52     \\
10000                         & -       & 2.93      & 9.32     & 23.86     \\ \hline
\end{tabular}
\end{sc}
\end{small}
\end{center}
\vskip -0.1in
\end{table}

\begin{table}[t!]
\centering
\caption{Test accuracy (\%) for \alg ($p=1$).}
\label{table:SAFL_Q}
\begin{small}
\begin{sc}
\begin{tabular}{cccccc}
\hline
\multirow{2}{*}{Dataset}  & \multirow{2}{*}{$s$} & \multicolumn{4}{c}{client update probability ($q$)}                                                           \\ \cline{3-6} 
                          &                    & \begin{tabular}[c]{@{}c@{}}1.0 \\ (FedAvg)\end{tabular} & 0.8   & 0.6   & 0.4   \\ \hline
\multirow{3}{*}{MNIST}    & 0                  & 84.49                                                   & 88.96 & 89.11 & 89.1  \\
                          & 2                  & 71.58                                                   & 88.11 & 88.06 & 88.27 \\
                          & 4                  & 57.05                                                   & 88.12 & 88.44 & 87.19 \\ \hline
\multirow{3}{*}{CIFAR-10} & 0                  & 75.7                                                    & 78.27 & 77.2  & 76.29 \\
                          & 2                  & 64.56                                                   & 77.13 & 75.17 & 75.08 \\
                          & 4                  & 50.7                                                    & 74.56 & 73.85 & 74.19 \\ \hline
\end{tabular}
\end{sc}
\end{small}
\vspace{-0.15in}
\end{table}

In Table~\ref{table:SAFL_vs_FedAvg_cifar10}, we show the comparison between our \alg algorithm and FedAvg algorithm on CIFAR-10 for incomplete client participation $s=4$. The observations in Section~\ref{sec: NumericalResults} are further illustrated: 1) There is non-negligible increase of the test accuracy for \alg algorithm with small amount of auxiliary data at server’s side. With 10000 data at server's side, the test accuracy increases by 23.86 \%. 2) There is actually no improvement with these auxiliary data for nearly homogeneous case (e.g., $p=10$), which is denoted by `-' in the table.


In Table~\ref{table:SAFL_Q}, we show the test accuracy of \alg on MNIST and CIFAR-10 under different client update probability $q$. 
Note that when $q=1.0$, \alg is equivalent to FedAvg. 
We can observe that even with a few server participation with a probability of $0.2$, there is a non-negligible improvement in test accuracy. 

\subsection{Fashion-MNIST}

We further run experiments with the Fashion-MNIST dataset, and the results are summarized as follows. These experiment results validate our theoretical findings.

\begin{table}[h]
\centering
\caption{Test accuracy (\%) for Fashion-MNIST dataset with different incomplete client participations.}
\begin{tabular}{|c|c|c|c|c|c|}
\hline
$s$ & 0 & 30 & 60 & 90 & 120 \\ \hline
Test accuracy & 87.71 $\pm$ 0.09 & 86.17 $\pm$ 1.57 & 82.37 $\pm$ 5.8 & 80.71 $\pm$ 0.38 & 78.53 $\pm$ 3.05 \\ \hline
\end{tabular}
\end{table}

\begin{table}[h]
\centering
\caption{Test accuracy (\%) for Fashion-MNIST dataset with different server's data. ($s=90, M=150, q=0.8$.)}
\begin{tabular}{|c|c|c|c|c|}
\hline
& FedAvg & SAFARI (1\%) & SAFARI (10\%) & SAFARI (20\%) \\ \hline
Test accuracy & 80 $\pm$ 0.38 & 82.0 $\pm$ 0.03 & 85.58 $\pm$ 1.05 & 85.14 $\pm$ 0.23 \\ \hline
\end{tabular}
\end{table}

\subsection{Ablation study about Random Initializations}

We have run each experiment setting with five random initializations for the MNIST dataset and three random initializations for the CIFAR10 dataset. We report the mean and standard variance. The results are summarized in the following tables, which will also be added in the revision.

\begin{table}[h]
\centering
\caption{Test accuracy (\%) for MNIST dataset. ($s$ is client sampling bias and $p$ is non-i.i.d. index.)}
\begin{tabular}{|c|c|c|c|c|c|}
\hline
 &  & $p=10$ & $p=5$ & $p=2$ & $p=1$ \\ \hline
$s=0$ & FedAvg & 92.67$\pm$ 0.05 & 89.70$\pm$ 0.23 & 86.04$\pm$ 0.61 & 84.60$\pm$ 0.99 \\ \cline{2-6}
 & SAFARI & 92.60$\pm$ 0.08 & 91.08$\pm$ 0.13 & 89.42$\pm$ 0.17 & 89.10$\pm$ 0.24 \\ \hline
$s=2$ & FedAvg & 92.64$\pm$ 0.07 & 89.13$\pm$ 0.28 & 86.34$\pm$ 0.92 & 71.66$\pm$ 0.74 \\ \cline{2-6}
 & SAFARI & 92.62$\pm$ 0.04 & 90.68$\pm$ 0.38 & 88.75$\pm$ 0.22 & 88.50$\pm$ 0.63 \\ \hline
$s=4$ & FedAvg & 92.58$\pm$ 0.03 & 88.76$\pm$ 0.13 & 77.82$\pm$ 0.22 & 57.09$\pm$ 0.04 \\ \cline{2-6}
 & SAFARI & 92.51$\pm$ 0.04 & 90.41$\pm$ 0.17 & 88.27$\pm$ 0.21 & 88.06$\pm$ 0.24 \\ \hline
\end{tabular}
\end{table}

\begin{table}[h]
\centering
\caption{Test accuracy (\%) for CIFAR10 dataset. ($s$ is client sampling bias and $p$ is non-i.i.d. index.)}
\begin{tabular}{|c|c|c|c|c|c|}
\hline
 &  & $p=10$ & $p=5$ & $p=2$ & $p=1$ \\ \hline
$s=0$ & FedAvg & 81.48$\pm$ 0.38 & 79.78$\pm$ 0.34 & 78.36$\pm$ 0.17 & 76.39$\pm$ 0.60 \\ \cline{2-6}
 & SAFARI & 81.40$\pm$ 0.45 & 80.13$\pm$ 0.33 & 79.00$\pm$ 0.25 & 78.34$\pm$ 0.12 \\ \hline
$s=2$ & FedAvg & 80.11$\pm$ 0.14 & 78.63$\pm$ 0.33 & 77.14$\pm$ 0.42 & 65.16$\pm$ 0.56 \\ \cline{2-6}
 & SAFARI & 80.41$\pm$ 0.10 & 79.13$\pm$ 0.28 & 77.66$\pm$ 0.10 & 77.21$\pm$ 0.39 \\ \hline
$s=4$ & FedAvg & 78.16$\pm$ 0.33 & 75.30$\pm$ 0.22 & 67.78$\pm$ 0.38 & 50.67$\pm$ 0.25 \\ \cline{2-6}
 & SAFARI & 79.28$\pm$ 0.21 & 78.27$\pm$ 0.21 & 76.52$\pm$ 0.16 & 75.18$\pm$ 0.54 \\ \hline
\end{tabular}
\end{table}

\subsection{Ablation Study about Number of Clients}

In addition to the important roles of $p$ and $s$, we agree that $m$ could serve as another influencing factor. Consequently, we have incorporated additional configurations for varying values of $m$, as delineated below. From these results, it is apparent that $m$ has a negative effect on test accuracy in non-i.i.d. scenarios. As the non-i.i.d. degree increases, the negative influence of $m$ becomes more pronounced.

\begin{table}[h]
\centering
\caption{Test accuracy (\%) for MNIST dataset. ($m$ is participated clients in training and $p$ is non-i.i.d. index.)}
\begin{tabular}{|c|c|c|c|c|c|}
\hline
 &  & $p=10$ & $p=5$ & $p=2$ & $p=1$ \\ \hline
$m=1$ & FedAvg & 92.11 & 65.63 & 58.65 & 49.86 \\ \cline{2-6}
 & SAFARI & 91.88 & 90.04 & 87.96 & 87.83 \\ \hline
$m=2$ & FedAvg & 92.40 & 83.28 & 67.03 & 53.75 \\ \cline{2-6}
 & SAFARI & 92.27 & 90.00 & 87.63 & 87.45 \\ \hline
$m=3$ & FedAvg & 92.51 & 87.19 & 75.14 & 55.76 \\ \cline{2-6}
 & SAFARI & 92.54 & 90.39 & 87.63 & 88.19 \\ \hline
$m=4$ & FedAvg & 92.55 & 88.60 & 78.52 & 56.42 \\ \cline{2-6}
 & SAFARI & 92.45 & 90.85 & 88.30 & 87.88 \\ \hline
$m=5$ & FedAvg & 92.62 & 88.81 & 77.81 & 57.05 \\ \cline{2-6}
 & SAFARI & 92.52 & 90.35 & 88.50 & 88.12 \\ \hline
$m=6$ & FedAvg & 92.58 & 88.20 & 77.42 & 57.12 \\ \cline{2-6}
 & SAFARI & 92.65 & 90.35 & 87.67 & 88.23 \\ \hline
\end{tabular}
\end{table}

\begin{table}[h]
\centering
\caption{Test accuracy (\%) for CIFAR10 dataset. ($m$ is participated clients in training and $p$ is non-i.i.d. index.)}
\begin{tabular}{|c|c|c|c|c|c|}
\hline
 &  & $p=10$ & $p=5$ & $p=2$ & $p=1$ \\ \hline
$m=1$ & FedAvg & 78.61 & 75.34 & 53.02 & 10.31 \\ \cline{2-6}
 & SAFARI & 79.80 & 78.70 & 75.90 & 63.04 \\ \hline
$m=2$ & FedAvg & 78.48 & 76.46 & 67.24 & 45.87 \\ \cline{2-6}
 & SAFARI & 79.15 & 77.90 & 76.75 & 72.93 \\ \hline
$m=4$ & FedAvg & 78.59 & 75.21 & 68.05 & 50.25 \\ \cline{2-6}
 & SAFARI & 79.98 & 77.56 & 76.60 & 75.51 \\ \hline
$m=5$ & FedAvg & 77.78 & 75.55 & 67.34 & 50.70 \\ \cline{2-6}
 & SAFARI & 79.47 & 78.48 & 76.66 & 74.56 \\ \hline
\end{tabular}
\end{table}


\end{document}